\documentclass[12pt]{article}

\usepackage{scicite}

\usepackage{hyperref} 
\usepackage{times}
\usepackage{amssymb}
\usepackage{amsmath}
\usepackage{algorithm}
\usepackage{algpseudocode}
\usepackage{setspace}
\let\Algorithm\algorithm
\renewcommand\algorithm[1][]{\Algorithm[#1]\setstretch{1.}}
\usepackage{graphicx}
\usepackage{rotating}
\usepackage{nopageno}
\usepackage{hyperref}

\usepackage{setspace}
\usepackage[font=singlespacing]{caption}

\usepackage[color=red]{todonotes}

\newenvironment{sciabstract}{%
\begin{quote} \bf}
{\end{quote}}

\newcommand{\nkey}{K^{\text{r}}}
\newcommand{\nm}{N^{\text{mem}}}
\newcommand{\ww}{v^{\text{wr}}}
\newcommand{\wre}{v^{\text{ret}}}
\newcommand{\sg}[1]{\text{StopGradient}({#1})}

\newcounter{lastnote}

\title{Unsupervised Predictive Memory in a \\ Goal-Directed Agent}

\author{{Greg Wayne$^{*,1}$, Chia-Chun Hung$^{*,1}$, David Amos$^{*,1}$,  Mehdi Mirza$^1$,} 
\\ {Arun Ahuja$^1$, Agnieszka Grabska-Barwi\'{n}ska$^1$, Jack Rae$^1$, Piotr Mirowski$^1$,}
\\ {Joel Z. Leibo$^1$, Adam Santoro$^1$, Mevlana Gemici$^1$, Malcolm Reynolds$^1$,}
\\ {Tim Harley$^1$, Josh Abramson$^1$, Shakir Mohamed$^1$, Danilo Rezende$^1$,}
\\ {David Saxton$^1$, Adam Cain$^1$, Chloe Hillier$^1$, David Silver$^1$,}
\\ {Koray Kavukcuoglu$^1$, Matt Botvinick$^1$, Demis Hassabis$^1$, Timothy Lillicrap$^1$.}
\\ \normalsize{$^{1}$DeepMind, 5 New Street Square, London EC4A 3TW, UK.}
\\
\normalsize{$^*$These authors contributed equally to this work.}
}

\date{}

\begin{document} 


\baselineskip24pt


\maketitle


\begin{sciabstract}
Animals execute goal-directed behaviours despite the limited range and scope of their sensors. To cope, they explore environments and store memories maintaining estimates of important information that is not presently available\cite{clayton1998episodic}. 
Recently, progress has been made with artificial intelligence (AI) agents that learn to perform tasks from sensory input, even at a human level, by merging reinforcement learning (RL) algorithms with deep neural networks\cite{mnih2015human, silver2016mastering}, and the excitement surrounding these results has led to the pursuit of related ideas as explanations of non-human animal learning\cite{song2017reward, hunt2017distributed}.
However, we demonstrate that contemporary RL algorithms struggle to solve simple tasks when enough information is concealed from the sensors of the agent, a property called ``partial observability''. 
An obvious requirement for handling partially observed tasks is access to extensive memory, but we show memory is not enough; it is critical that the right information be stored in the right format. We develop a model, the \emph{Memory, RL, and Inference Network} (MERLIN), in which memory formation is guided by a process of predictive modeling. MERLIN facilitates the solution of tasks in 3D virtual reality environments\cite{beattie2016deepmind} for which partial observability is severe and memories must be maintained over long durations. Our model demonstrates a single learning agent architecture that can solve canonical behavioural tasks in psychology and neurobiology without strong simplifying assumptions about the dimensionality of sensory input or the duration of experiences.
\end{sciabstract}

\section*{Introduction}

Artificial intelligence research is undergoing a renaissance as RL techniques\cite{sutton1998reinforcement}, which address the problem of optimising sequential decisions, have been combined with deep neural networks into artificial agents that can make optimal decisions by processing complex sensory data\cite{mnih2015human}. In tandem, new deep network structures have been developed that encode important prior knowledge for learning problems. One important innovation has been the development of neural networks with external memory systems, allowing computations to be learned that synthesise information from a large number of historical events\cite{weston2014memory,bahdanau2014neural,graves2016hybrid}. 

Within RL agents, neural networks with external memory systems have been optimised ``end-to-end'' to maximise the amount of reward acquired during interaction in the task environment. That is, the systems learn how to select relevant information from input (sensory) data, store it in memory, and read out relevant memory items purely from trial-and-error action choices that led to higher than expected reward on tasks. While this approach to artificial memory has led to successes\cite{todd2009learning, oh2016control,graves2016hybrid,duan2017one}, we show that it fails to solve simple tasks drawn from behavioural research in psychology and neuroscience, especially ones involving long delays between relevant stimuli and later decisions: these include, but are not restricted to, problems of navigation back to previously visited goals\cite{tolman1948cognitive, tse2007schemas, o1978hippocampus}, rapid reward valuation\cite{corbit2000role}, where an agent must understand the value of different objects after few exposures, and latent learning, where an agent acquires unexpressed knowledge of the environment before being probed with a specific task\cite{blodgett1929effect, tolman1930introduction}.

We propose MERLIN, an integrated AI agent architecture that acts in partially observed virtual reality environments and stores information in memory based on different principles from existing end-to-end AI systems: it learns to process high-dimensional sensory streams, compress and store them, and recall events with less dependence on task reward. We bring together ingredients from external memory systems, reinforcement learning, and state estimation (inference) models and combine them into a unified system using inspiration from three ideas originating in psychology and neuroscience: predictive sensory coding\cite{rao1999predictive, bastos2012canonical, hindy2016linking}, the hippocampal representation theory of Gluck and Myers\cite{gluck1993hippocampal,moustafa2013trace}, and the temporal context model and successor representation\cite{howard2002distributed,dayan1993improving,stachenfeld2017hippocampus}. To test MERLIN, we expose it to a set of canonical tasks from psychology and neuroscience, showing that it is able to find solutions to problems that pose severe challenges to existing AI agents. MERLIN points a way beyond the limitations of end-to-end RL toward future studies of memory in computational agents. 


RL formalises the problem of finding a \emph{policy} $\pi$ or a mapping from sensory observations $o$ to actions $a$. A leading approach to RL begins by considering policies that are \emph{stochastic}, so that the policy describes a distribution over actions. Memory-free RL policies that directly map instantaneous sensory data to actions fail in partially observed environments where the sensory data are incomplete. Therefore, in this work we restrict our attention to \emph{memory-dependent} policies, where the action distribution depends on the entire sequence of past observations. 

\begin{figure}
\centering 
\includegraphics[width=1\textwidth]{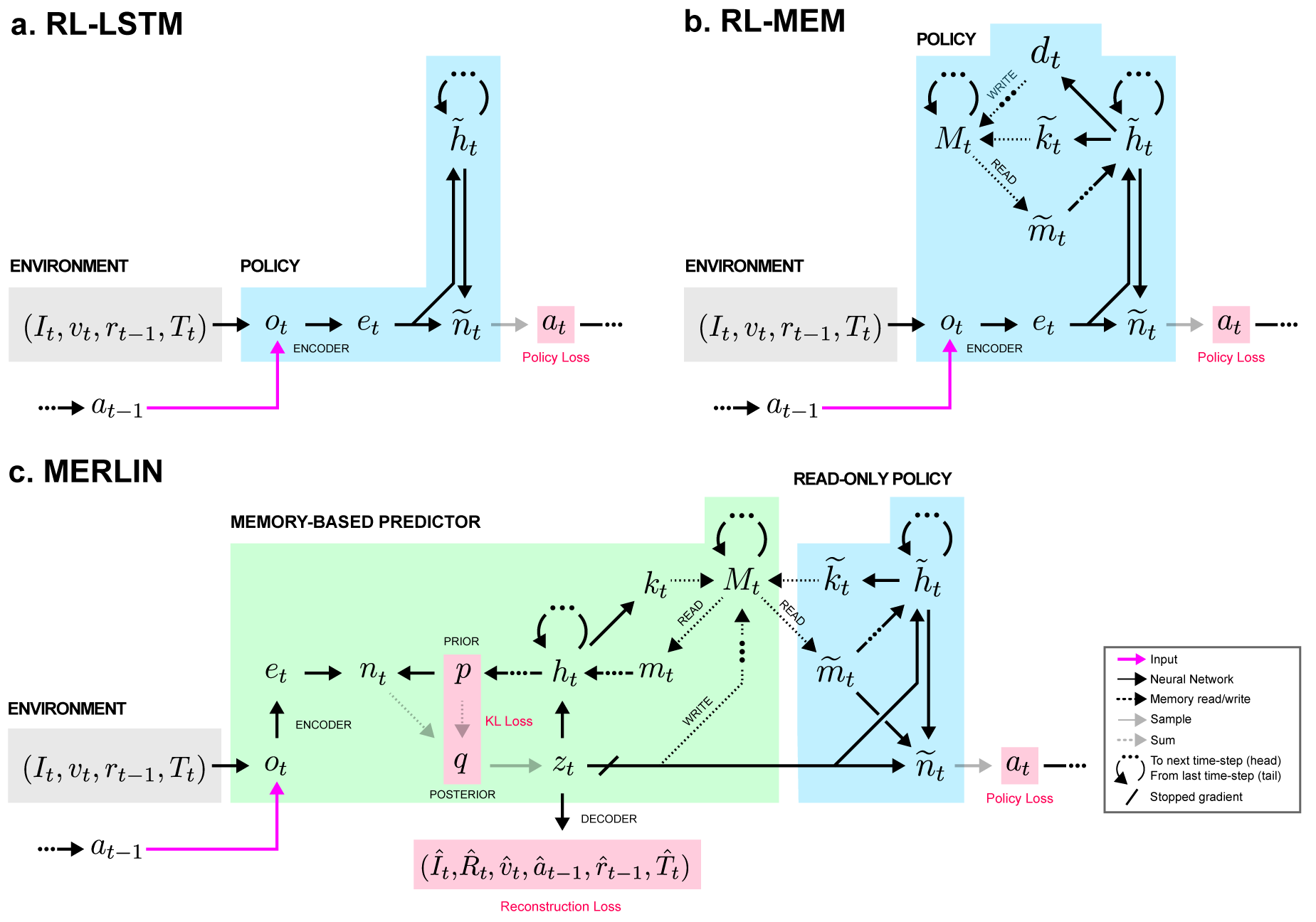}
\caption{\footnotesize{\textbf{Agent Models}. \textbf{a}. At time $t$, RL-LSTM receives the environment sensory input $o_t$ composed of the image $I_t$, self-motion velocity $v_t$, the previous reward $r_{t-1}$, and the text input $T_t$. These are sent through an encoder, consisting of several modality-specific encoders to make an embedding vector $e_t$. This is provided as input to a recurrent LSTM network, $\tilde{h}_t$, which outputs through a neural network with intermediate hidden layer $\tilde{n}_t$ the action probabilities. An action $a_t$ is sampled and acts on the environment. The whole system is optimised to maximise the sum of future rewards via the policy loss. \textbf{b}. RL-MEM is similar except that the recurrent $\tilde{h}_t$ reads from a memory matrix $M_t$ using several \emph{read heads} that each produces a key vector $\tilde{k}_t$ that is compared by vector similarity (normalised dot product) to the rows of the memory. The most similar rows are averaged and returned as read vectors $\tilde{m}_t$ that are all concatenated. This read is provided as input to the recurrent network at time $t+1$ and influences the action probabilities at the current time. The recurrent network has an additional output write vector $d_t$ that is inserted to row $t$ of the memory at time $t+1$. The RL-LSTM and RL-MEM architectures also learn a return prediction as a network connected to the policy as in standard A3C\cite{mnih2016asynchronous}. These are suppressed here for simplicity but discussed in Methods~Sec.~5.1.
\textbf{c}. Sensory input in MERLIN flows first through the MBP, whose recurrent network $h$ has produced a prior $p$ distribution over the state variable $z_t$ at the previous time step $t-1$. The mean and log standard deviation of the Gaussian distribution $p$ are concatenated with the embedding and passed through a network to form an intermediate variable $n_t$, which is added to the prior to make a Gaussian posterior distribution $q$, from which the state variable $z_t$ is sampled. This is inserted into row $t$ of the memory matrix and passed to the recurrent network $h_t$ of the memory-based predictor (MBP). This recurrent network has several read heads each with a key $k_t$, which is used to find matching items $m_t$ in memory. The state variable is passed as input to the read-only policy and is passed through decoders that produce reconstructed input data (with carets) and the Gluck and Myers\cite{gluck1993hippocampal} return prediction $\hat{R}_t$. The MBP is trained based on the VLB objective\cite{kingma2013auto, rezende2014stochastic}, consisting of a reconstruction loss and a KL divergence between $p$ and $q$. To emphasise the independence of the policy from the MBP, we have blocked the gradient from the policy loss into the MBP.}}
\label{fig:architectures}
\end{figure}

In Fig.~\ref{fig:architectures}a, we see a standard memory-dependent RL policy architecture, RL-LSTM, which is a well-tuned variant of the ``Advantage Actor Critic'' architecture (A3C)\cite{mnih2016asynchronous} with a deeper convolutional network visual encoder. At each time $t$, a sensory encoder network takes in an observation $o_t$ and produces an embedding vector $e_t = \text{enc}(o_t)$. This is passed to a recurrent neural network\cite{hochreiter1997long}, which has a memory state $h_t$ that is produced as a function of the input and the previous state: $h_{t} = \text{rec}(h_{t-1}, e_t)$. Finally, a probability distribution indicating the probability of an action is produced as a function of the memory state $\Pr(a_t) = \pi( a_t | h_t)$. The encoder, recurrent network, and action distribution are all understood to be neural networks with optimisable parameters $\theta=[\theta_{\text{enc}};\theta_{\text{rec}};\theta_{\pi}]$. An agent with relatively unstructured recurrence like RL-LSTM can perform well in partially observed environments but can fail to train when the amount of information that must be recalled is sufficiently large. 

In Fig.~\ref{fig:architectures}b, we see an RL agent (RL-MEM) augmented with an external memory system that stores information in a matrix $M$. In addition to the state of the recurrent network, the external memory stores a potentially larger amount of information that can be read from using a \emph{read key} $k_t$, which is a linear function of $h_t$ that is compared against the contents of memory: $m_t = \text{read}(M_{t}, k_t)$. The recurrent network is updated based on this read information with $h_{t} = \text{rec}(h_{t-1}, m_t, e_t)$. The memory is written to at each time step by inserting a vector $d_t$, produced as a linear function of $h_t$, into an empty row of memory: $M_{t+1} = \text{write}(M_{t}, d_t)$. The functions ``$\text{read}$'' and ``$\text{write}$'' additionally have parameters so that $\theta=[\theta_{\text{enc}};\theta_{\text{rec}};\theta_{\pi};\theta_{\text{read}};\theta_{\text{write}}]$. 

An agent with a large external memory store can perform better on a range of tasks, but training perceptual representations for storage in memory by end-to-end RL can fail if a task demands high-fidelity perceptual memory. In RL-LSTM/MEM, the entire system, including the representations formed by the encoder, the computations performed by the recurrent network, the rules for reading information from memory and writing to it (for RL-MEM), and the action distribution are optimised to make trial and error actions more or less likely based on the amount of reward received. RL thus learns to encode and retrieve information based on trial and error decisions and resultant rewards. This is indirect and inefficient: sensory data can instead be encoded and stored without trial and error in a temporally local manner.

Denoting the reward at time $t$ as $r_t$, the aim is to maximise the sum of rewards that the policy receives up to a final time, known as the \emph{return} $R_t = r_t + r_{t+1} + r_{t+2} + \dots + r_T$. This is achieved by a ``policy gradient'' update rule\cite{sutton2000policy} that increases the log probability of the selected actions based on the return (see Methods~Sec.~4.4):
\begin{equation}
\Delta \theta \propto \sum_{t=0}^{T} R_t \nabla_\theta \log \pi_\theta(a_t | h_t). \label{eq:policy_gradient}
\end{equation}
In practice, Eq.~\ref{eq:policy_gradient} is implemented by the truncated backpropagation-through-time algorithm\cite{werbos1990backpropagation,sutskever2013training} over a fixed window $\tau$ defining the number of time steps over which the gradient is calculated\cite{mnih2016asynchronous}. Intuitively, it defines the duration over which the model can assign credit or blame to network dynamics or information storage events leading to success or failure. When $\tau$ is smaller than the typical time scale over which information needs to be stored and retrieved, RL models can struggle to learn at all. Thus, learning the representations to put in memory by end-to-end policy gradient RL only works if the minimal time delay between encoding events and actions is not too long -- for example, larger than the window $\tau$.

MERLIN (Fig.~\ref{fig:architectures}c) optimises its representations and learns to store information in memory based on a different principle, that of unsupervised prediction\cite{rao1999predictive,bastos2012canonical,hindy2016linking}. MERLIN has two basic components: a \emph{memory-based predictor} (MBP) and a policy. The MBP is responsible for compressing observations into low-dimensional state representations $z$, which we call \emph{state variables}, and storing them in memory. The state variables in memory in turn are used by the MBP to make predictions guided by past observations. This is the key thesis driving our development: an agent's perceptual system should produce compressed representations of the environment; predictive modeling is a good way to build those representations; and the agent's memory should then store them directly. The policy can primarily be the downstream recipient of those state variables and memory contents. 

Machine learning and neuroscience have both engaged with the idea of unsupervised and predictive modeling over several decades\cite{kok2012less}. Recent discussions have proposed such predictive modeling is intertwined with hippocampal memory\cite{hindy2016linking, finkelstein20163}, allowing prediction of events using previously stored observations, for example, of previously visited landmarks during navigation or the reward value of previously consumed food. MERLIN is a particular and pragmatic instantiation of this idea that functions to solve challenging partially observable tasks grounded in raw sensory data.

We combine ideas from state estimation and inference\cite{kalman1960new} with the convenient modern framework for unsupervised modeling given by variational autoencoders\cite{kingma2013auto, rezende2014stochastic, chung2015recurrent, gemici2017generative} as the basis of the MBP (Fig.~\ref{fig:architectures}c). Information from multiple modalities (image $I_t$, egocentric velocity $v_t$, previous reward $r_{t-1}$ and action $a_{t-1}$, and possibly a text instruction $T_t$) constitute the MBP's observation input $o_t$ and are encoded to $e_t = \text{enc}(o_t)$. A probability distribution, known as the \emph{prior}, predicts the next state variable conditioned on a history maintained in memory of the previous state variables and actions: $p(z_{t} | z_1, a_1, \dots, z_{t-1}, a_{t-1})$. Another probability distribution, the \emph{posterior}, corrects this prior based on the new observations $o_t$ to form a better estimate of the state variable: $q(z_t | z_1, a_1, \dots, z_{t-1}, a_{t-1}, o_t)$. The posterior samples from a Gaussian distribution a realisation of the state variable $z_t$, and this selected state variable is provided to the policy and stored in memory. In MERLIN, the policy, which has read-only access to the memory, is the only part of the system that is trained conventionally according to Eq.~\ref{eq:policy_gradient}.

The MBP is optimised to function as a ``world model''\cite{neisser1967cognitive, barlow1987cerebral}: in particular, to produce predictions that are consistent with the probabilities of observed sensory sequences from the environment: $\Pr(o_1,o_2,\dots)$. This objective can be intractable, so the MBP is trained instead to optimise the variational lower bound (VLB) loss, which acts as a tractable surrogate. One term of the VLB is reconstruction of observed input data. To implement this term, several decoder networks take $z_t$ as input, and each one transforms back into the space of a sensory modality ($\hat{I}_t$ reconstructs image $I_t$; the others are self-motion $\hat{v}_t$, text input $\hat{T}_t$, previous action $\hat{a}_{t-1}$, and previous reward $\hat{r}_{t-1}$). The difference between the decoder outputs and the ground truth data is the loss term. The VLB also has a term that penalises the difference (KL divergence) between the prior and posterior probability distributions, which ensures that the predictive prior is consistent with the posterior produced after observing new stimuli.

Although it is desirable for the state variables to faithfully represent perceptual data, we still want them to emphasise, when possible, rewarding elements of the environment over and above irrelevant ones. To accomplish this, we follow the hippocampal representation theory of Gluck and Myers\cite{gluck1993hippocampal}, who proposed, as an account of diverse phenomena in animal conditioning, that hippocampal representations pass through a compressive bottleneck and then reconstruct input stimuli together with task reward. In our context, $z_t$ is the compressive bottleneck, and we include an additional decoder that makes a prediction $\hat{R}_t$ of the return $R_t$ as a function of $z_t$. 
Algorithms such as A3C predict task returns and use these predictions to reduce variance in policy gradient learning\cite{mnih2016asynchronous}. In MERLIN, return prediction also has the essential role of shaping state representations constructed by unsupervised prediction.
Including this prediction has an important effect on the performance of the system, encouraging $z_t$ to focus on compressing sensory information while maintaining information of significance to tasks. In the 3D virtual reality environments described subsequently, sensory input to the agent comprises order $10^4$ dimensions, whereas the state variable is reduced to order $10^2$; this is achieved without losing information critical to task-related computations.

The utility of the memory system can be further increased by storing each state variable together with a representation of the events that occurred after it in time, which we call \emph{retroactive memory updating}. In navigation, for example, this allows perceptual information related to a way-point landmark to be stored next to information about a subsequently experienced goal. We implement this by an edit of the memory matrix in which a filtered sum of state variables produced after $z_t$ is concatenated in the same row: $[z_t, (1 - \gamma) \sum_{t' > t} \gamma^{t'-t} z_{t'}]$, with $\gamma < 1$. 

Further details about MERLIN are available in Methods.

\begin{figure}
\centering 
\includegraphics[width=1\textwidth]{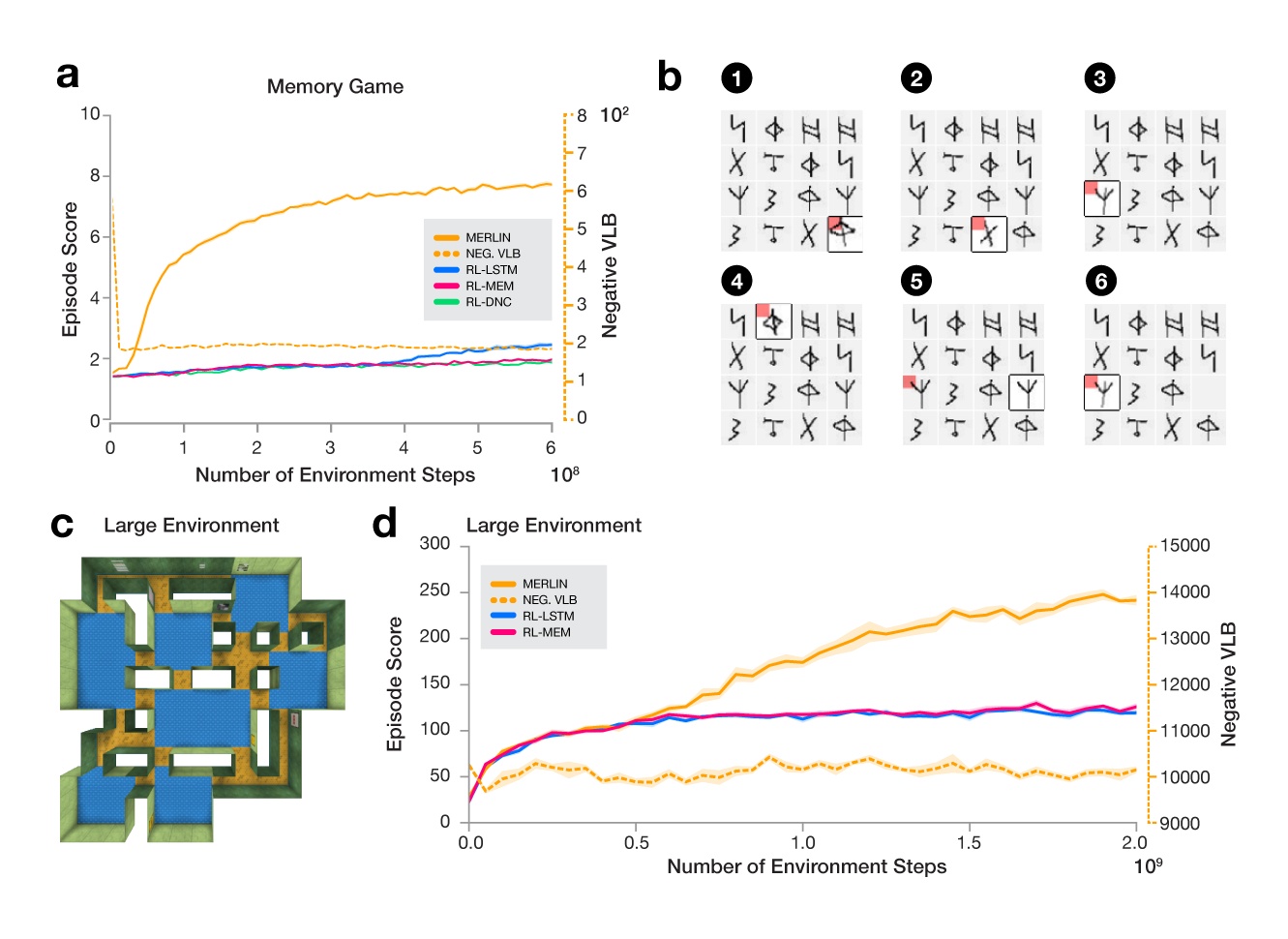}
\caption{\footnotesize{\textbf{Basic Tasks}. \textbf{a}. On the Memory Game, MERLIN exhibited qualitatively superior learning performance compared to RL-LSTM and RL-MEM (yellow: average episode score as a function of the number of steps taken by the simulator; hashed line: the cost of the MBP -- the negative of the variational lower bound). We include an additional comparison to a Differentiable Neural Computer (DNC)-based agent as well\cite{graves2016hybrid}, which we did not study further as its computational complexity scales quadratically with time steps if no approximations are made. ``Number of Environment Steps'' is the total number of forward integration steps computed by the simulator. (The standard error over five trained agents per architecture type was nearly invisibly small.) \textbf{b}. MERLIN playing the memory game: an omniglot image observation\cite{lake2015human} (highlighted in white) was jointly coded with its location, which was given by the previous action, on the grid. MERLIN focused its memory access on a row in memory storing information encoded when it had flipped the card indicated by the red square. On step 5, MERLIN retrieved information about the historical observation that was most similar to the currently viewed observation. On step 6, MERLIN chose the previously seen card and scored a point. \textbf{c}. A randomly generated layout for a large environment. The Large Environment Task was one of the four navigation tasks on which MERLIN was simultaneously trained. The agent sought a fixed goal (worth 10 points of reward) as many times as possible in 90 seconds, having been teleported to a random start location after each goal attainment. \textbf{d}. Learning curves for the Large Environment showed that MERLIN revisited the goal more than twice as often per episode as comparison agents. ``Number of Environment Steps'' logged the number of environment interactions that occurred across all four tasks in the set.}}
\label{fig:memory_and_nav}
\end{figure}

We first consider a very simple task that RL agents should easily solve, the children's game ``Memory'', which has even been comparatively studied in rhesus monkeys and humans\cite{washburn2002species}. 
Here, cards are placed face down on a grid, and one card at a time is flipped over then flipped back. 
If two sequentially flipped cards show matching pictures, a point is scored, and the cards removed. 
An agent starts from a state of ignorance and must explore by flipping cards and remembering the observations. Strikingly, RL-LSTM and MEM were unable to solve this task, whereas MERLIN found an optimal strategy (Fig.~\ref{fig:memory_and_nav}a). A playout by MERLIN is shown in Fig.~\ref{fig:memory_and_nav}b, in which it is seen that MERLIN read from a memory row storing information about the previously observed matching card one time step before MERLIN's policy flipped that card.

\begin{figure}
\centering 
\vspace{-1.4cm}
\includegraphics[width=1\textwidth]{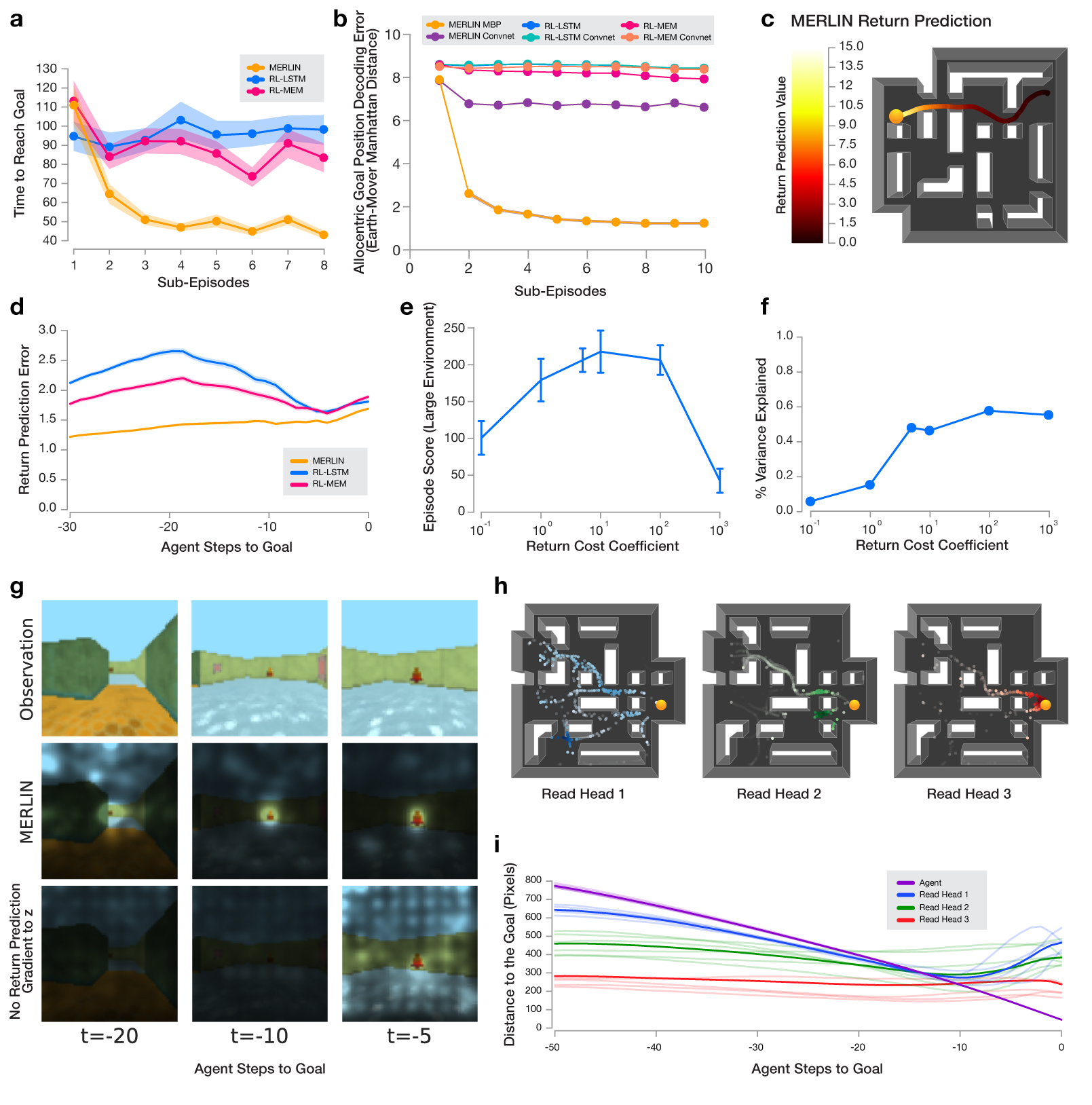} 
\vspace{-1.2cm}
\caption{\footnotesize{\textbf{Analysis}. \textbf{a}. MERLIN alone was able to relocate the goal in less time on subsequent visits. \textbf{b}. Each large environment was laid out on a $15 \times 15$ unit spatial grid. From $[z_t, h_t, m_t]$, the allocentric $(x,y)$ position of the goal was decoded to an average accuracy of $1.67$ units in Manhattan distance (subgrid accuracy per dimension). This accuracy improved with sub-episodes, implying MERLIN integrated information across multiple transits to goal. (Ext.~Fig.~1 has an egocentric decoding.) Decoding from feedforward visual convnets (e.g. MERLIN Convnet), from the policy $h_t$ in RL-LSTM, or from $[h_t, m_t]$ in RL-MEM did not yield equivalent localisation accuracy. \textbf{c}. An example trajectory while returning to the goal (orange circle): even when the goal was not visible, MERLIN's return prediction climbed as it anticipated the goal. \textbf{d}. MERLIN's return prediction error was low on approach to the goal as the MBP used memory to estimate proximity. \textbf{e}. Task performance varied with different coefficients of the return prediction cost with a flat maximum that balanced sensory prediction against return prediction. \textbf{f}. For higher values of the coefficient, regression using a feedforward network from the return prediction $\hat{R}_t$ to $z_t$ explained increasingly variance in $z_t$. Thus, the state variables devoted more effective dimensions to code for future reward. \textbf{g}. \emph{Top row}: observations at 20, 10, and 5 steps before goal attainment. \emph{Middle}: the L2 norm of the gradient of the return prediction with respect to each image pixel, $\sum_{c=1}^3 (\partial \hat{R}_t / \partial I_t^{w,h,c})^2$ ($c$ for colour channel), was used to intensity mask the observation, automatically segmenting pixels containing the goal object. \emph{Bottom}: a model trained without a gradient from the return prediction through $z$ was not similarly sensitive to the goal. (The gradient pathway was unstopped for analysis.) \textbf{h}. The three memory read heads of the MBP specialised to focus on, respectively, memories formed at locations ahead of the agent, at waypoint doorways \emph{en route} to the goal, and around the goal itself. \textbf{i}. Across a collection of trained agents, we assigned each MBP read head the index 1, 2, or 3 if it read from memories formed far from (blue), midway to (green), or close to the goal (red). The reading strategy in panel 3h with read heads specialised to recall information formed at different distances from the goal developed reliably.}}
\label{fig:nav}
\end{figure}

MERLIN also excels at solving one-shot navigation problems from raw sensory input in randomly generated, partially observed 3D environments. We trained agents to be able to perform on any of four variants of a goal-finding navigation task (Ext.~Video~1: \url{https://youtu.be/YFx-D4eEs5A}). These tested the ability to locate a goal in a novel environment map and quickly return to it. After reaching the goal, the agent was teleported to a random starting point. It was allowed a fixed time on each map and rewarded for each goal visit. To be successful, an agent had to rapidly build an approximate model of the map from which it could navigate back to the goal.  

The base task took place in environments with 3-5 rooms. The variations included a task where doors dynamically opened and closed after reaching the goal, inspired by Tolman\cite{tolman1948cognitive}; a task where the skyline was removed to force navigation based on proximal cues; the same task in larger environments with twice the area and a maximum of 7 rooms (Fig.~\ref{fig:memory_and_nav}c).
MERLIN learned faster and reached higher performance than comparison agents and professional human testers (Ext.~Table~1; Fig.~\ref{fig:memory_and_nav}d; Ext.~Fig.~2). MERLIN exhibited robust memory-dependent behaviour as it returned to the goal in less time on each repeated visit, having rapidly apprehended the layout of the environment (Fig.~\ref{fig:nav}a).
Within very few goal attainments in each episode in large environments, it was possible to classify the absolute position of the goal to high accuracy from the MBP state (state variable $z_t$, memory reads $m_t$, and recurrent state $h_t$) (Fig.~\ref{fig:nav}b), demonstrating that MERLIN quickly formed allocentric representations of the goal location. 

Even when a goal was out of view, MERLIN's return predictions rose in expectation of the oncoming goal (Fig.~\ref{fig:nav}c), and its return prediction error was lower than the analogous value function predictions of RL-LSTM and RL-MEM (Fig.~\ref{fig:nav}d). Agent performance was robust to a range of different weights on the return prediction cost coefficient, but for very low and high values, performance was dramatically affected, as the state variables became insensitive to reward for low values and insensitive to other sensors for high values (Fig.~\ref{fig:nav}e). Decoding the MBP prior distribution's mean over the next state variable could be used to check visual accuracy across a spectrum of weights on the return prediction cost coefficient; lower values produced cleaner visual images, retaining more information (Ext.~Fig.~3). Regressing from the return predictions $\hat{R}_t$ to $z_t$ showed that the return prediction explained more variance of the state variable for higher return cost coefficients. We also observed the emergent phenomenon that the region of the visual field to which the return prediction was sensitive was a segmentation around the goal (Fig.~\ref{fig:nav}g). An agent trained to predict return but whose prediction errors did not send gradients during training through the state variable did not develop these receptive fields (Fig.~\ref{fig:nav}g).

Remarkably, though it had not been explicitly programmed, MERLIN showed evidence of hierarchical goal-directed behaviour, which we detected from the MBP's read operations. 
The three read heads of the MBP specialised to perform three functions. One would read from memories associated with previously visited locations just ahead of the agent's movement (Fig.~\ref{fig:nav}h, left); a second from memories associated with the goal location (Fig.~\ref{fig:nav}h, right); intriguingly, a third alternated between memories associated with various important sub-goals -- particularly doorways near the room containing the goal (Fig.~\ref{fig:nav}h~center). 
Across a group of 5 trained agents, this specialisation of heads attending to memories formed at different distances from the goal emerged robustly (Fig.~\ref{fig:nav}i).

\begin{figure}
\centering 
\includegraphics[width=\textwidth]{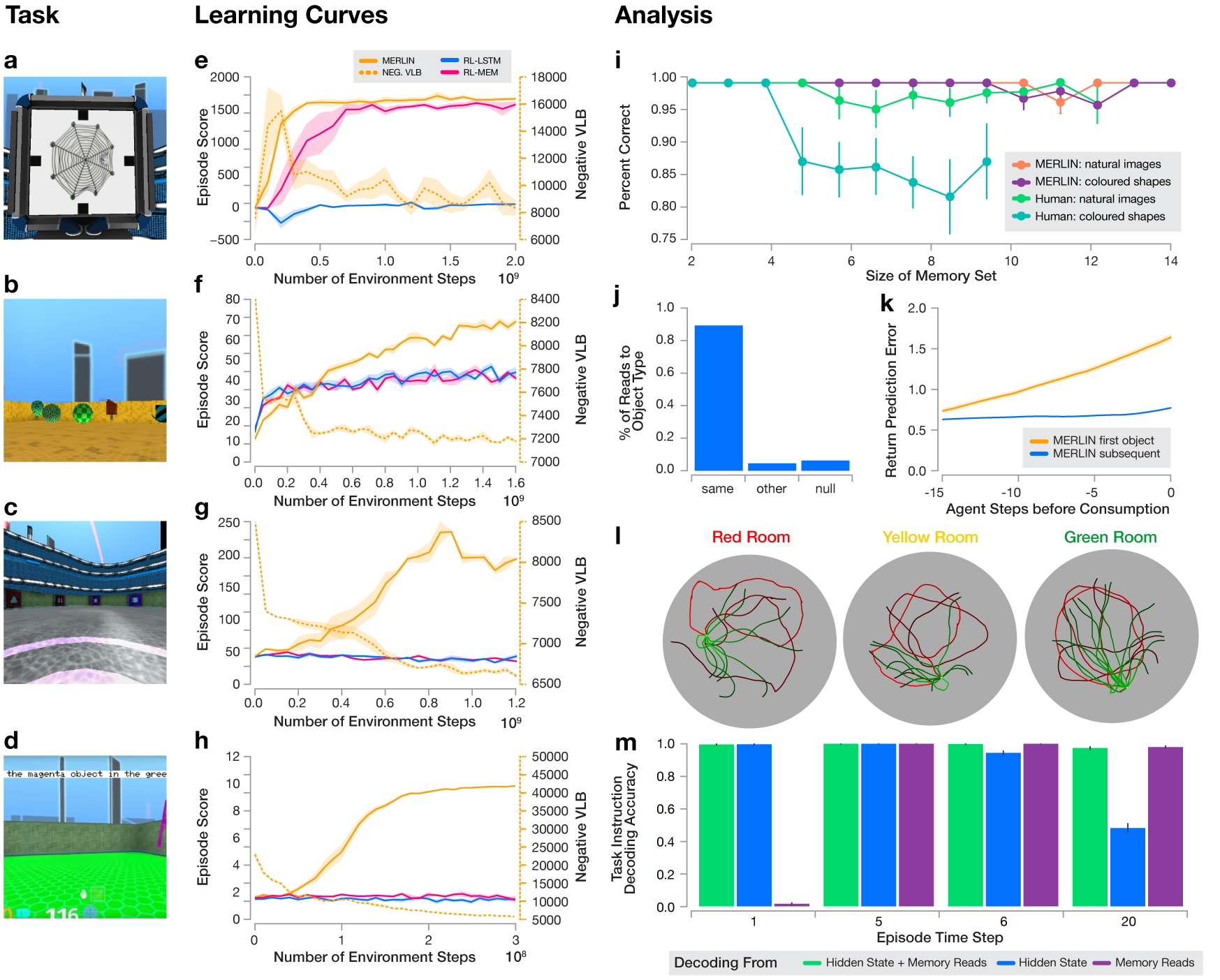}
\caption{\footnotesize{\textbf{Task Battery}. \textbf{a}. Arbitrary Visuomotor Mapping. The agent had to remember the target direction associated with each image. \textbf{b}. Rapid Reward Valuation. The agent had to explore to find good objects and thenceforth consume only those. \textbf{c}. Episodic Water Mazes.  The agent experienced one of three water mazes identifiable by wall colour and had to relocate the platform location. \textbf{d}. Executing Transient Instructions. The agent was given a verbal instruction indicating the room colour and object colour to find. \textbf{e-h}. Learning Curves for Corresponding Tasks. The dotted yellow lines represent the negative of the VLB of the MBP. \textbf{i}. In comparison to a human tester, MERLIN exhibited better accuracy as a function of the number of items to recall. On a new set of synthetic images made of constellations of coloured shapes and letters (Ext.~Fig.~4), MERLIN retained higher performance. \textbf{j}. The MBP read memories of the same object as the one about to be consumed. \textbf{k}. MERLIN's return predictions became more accurate after the first experience consuming each object type. \textbf{l}. MERLIN developed an effective exploration strategy (red) and on subsequent visits to each of three rooms exhibited directed paths to remembered platforms (green). \textbf{m}. The MBP variables $h_t, m_t$ were sufficient to reconstruct the instruction on test data at diverse times in the episode, even when the instruction was no longer present.}}
\label{fig:battery}
\end{figure}

To demonstrate its generality, we applied MERLIN to a battery of additional tasks. The first, ``Arbitrary Visuomotor Mapping'', came from the primate visual memory literature\cite{wise2000arbitrary} and demanded learning motor response associations to complex images\cite{wang2016learning}. The agent needed to fixate on a screen and associate each presented image with movement to one of four directions (Fig.~\ref{fig:battery}a; Ext.~Video~2: \url{https://youtu.be/IiR_NOomcpk}). At first presentation, the answer was indicated with a colour cue but subsequently needed to be recalled. With correct answers, the number of images to remember was gradually increased. MERLIN solved the task essentially without error, reaching performance above human level (Fig.~\ref{fig:battery}e\&i). When probed on a set of multi-object synthetic images modeled on a visual working memory task\cite{luck1997capacity} (Ext.~Fig.~4), MERLIN generalised immediately, exhibiting accuracy declines at higher memory loads than a human subject (Fig.~\ref{fig:battery}i). This transfer result implied that MERLIN learned the task structure largely independently of the image set. Moreover, MERLIN was able to learn, exclusively through unsupervised mechanisms, to distinguish complex images -- even previously unseen ones with different statistics.

In a second task, MERLIN demonstrated the ability to perform rapid reward valuation, a facility subserved by the hippocampus\cite{corbit2000role}. The agent was placed in a room with a selection of eight kinds of objects from a large pool with random textures that it could consume (Fig.~\ref{fig:battery}b; Ext.~Video~3: \url{https://youtu.be/dQMKJtLScmk}). Each object type was assigned a positive or negative reward value at the beginning of the episode; after clearing all positive reward from the environment, the agent started a second sub-episode with the same objects and had a chance to consume more. 
MERLIN learned to quickly probe and retain knowledge of object value. When approaching an object, it focused its reads on memories of the same object in preference to memories of others (Fig.~\ref{fig:battery}j), suggesting that it had formed view and background-invariant representations. MERLIN used these reads to make predictions of the upcoming value of previously consumed objects (Fig.~\ref{fig:battery}k). Here, the retroactive memory updates were very effective: the retroactive portion of a memory row written the first time an object was exploratively approached could be used to decode the reward value of the subsequent consumption  with $93\%$ accuracy ($N=25,000$, 5-fold cross-validation: Methods Sec.~9.7).

We next probed MERLIN's ability to contextually load episodic memories and retain them for a long duration. In one task, over six minutes MERLIN experienced three differently coloured ``water mazes''\cite{morris1984developments} with visual cues on the walls (Fig.~\ref{fig:battery}c) and each one with a randomly-positioned hidden platform as a goal location. MERLIN learned to explore the mazes and to store relevant information about them in memory and retrieve it without interference to relocate the platforms from any starting position (Fig.~\ref{fig:battery}l; Ext.~Fig.~6; Ext.~Video~4: \url{https://youtu.be/xrYDlTXyC6Q}). 

MERLIN learned to respond correctly to transiently communicated symbolic commands\cite{pilley2011border, hermann2017grounded}. For the first five steps in an episode, an instruction to retrieve the ``$<$colour$>$ object from the $<$colour$>$ room'' was presented to the text encoding network (Fig.~\ref{fig:battery}d; Ext.~Video~5: \url{https://youtu.be/04H28-qA3f8}). By training feedforward network classifiers with input the recurrent state $h_t$ and memory readouts $m_t$ of the MBP, we found it was possible to decode the instruction text tokens on held-out data at any point in the episode (Fig.~\ref{fig:battery}m), demonstrating persistent activity encoding task-related representations.

\begin{figure}
\centering 
\includegraphics[width=\textwidth]{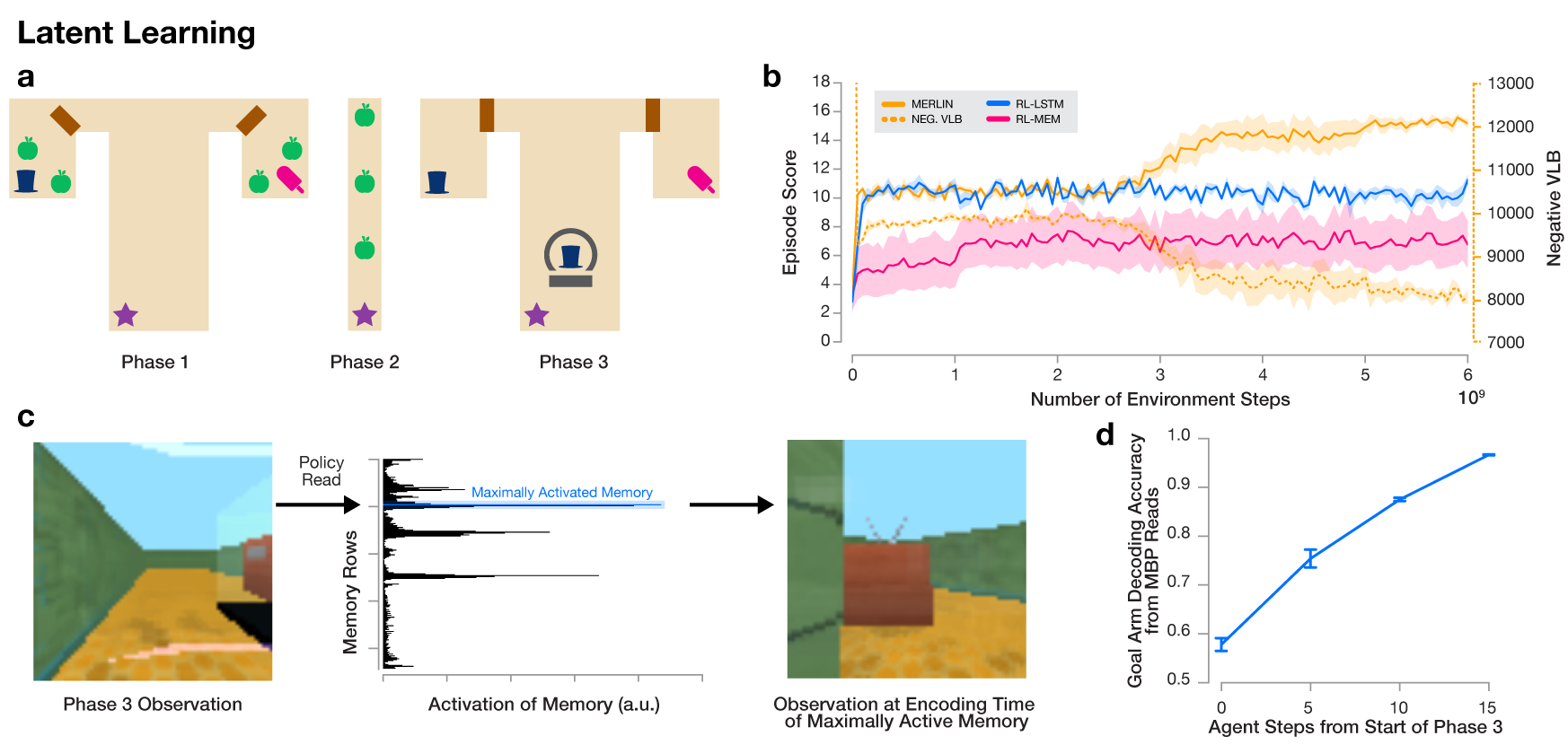}
\caption{\footnotesize{\textbf{Latent Learning}. \textbf{a}. \emph{Left}: Phase 1: the agent, beginning at the purple starred location, had to engage in an apple collecting task in a T-Maze with open doors and a random object in each room, initially valueless. \emph{Middle}: Phase 2: after collecting the apples in phase 1, the agent engaged in a distractor task to collect apples in a long corridor. \emph{Right}: Phase 3: a glass box containing one of the same objects cued the agent to find that object in one of the rooms behind a closed door. The agents only had one chance to choose. \textbf{b}. MERLIN was able to learn the task while the end-to-end agents performed at random in phase 3. \textbf{c}. After training, at the beginning of phase 3, the policy's read could be seen to recall maximally a memory stored in phase 1 while approaching the same object. \textbf{d}. The MBP reads also contained information about the location of the cued object. A logistic regression classifier could be trained with nearly perfect accuracy to predict the goal arm from the MBP reads on 5-fold cross-validated data ($N=1,000$).}}
\label{fig:latent}
\end{figure}

Finally, we examined a latent learning effect\cite{seward1949experimental, thistlethwaite1951critical}. In a T-Maze environment (Fig.~\ref{fig:latent}a), the agent was first given an apple collecting task (phase 1). Random objects were placed in the left and right maze arm rooms, which were occluded from outside the rooms. After collecting the apples in phase 1, the agent engaged in an unrelated distractor task of further apple collecting in a long corridor (phase 2). In phase 3, the agent was teleported to the beginning of the T-Maze where a glass box contained one of the two objects in the arm rooms. With the doors to the rooms now closed and only one chance to choose a room, MERLIN learned to find the object corresponding to the container cue (Ext.~Video~6: \url{https://youtu.be/3iA19h0Vvq0}), demonstrating that it could predicate decisions on observational information acquired before the stimuli were associated to the probe task (Fig.~\ref{fig:latent}b), even though the phase 2 delay period was longer than the backpropagation-through-time window $\tau$. On observation of the container, the read-only policy could recall the observation of the object made during phase 1 (Fig.~\ref{fig:latent}c). The MBP reads $m_t$ could also be used at the same time to decode the correct arm containing the cued object (Fig.~\ref{fig:latent}d).

As we have intimated, the architecture in Fig.~\ref{fig:architectures}c is not the unique instantiation of the principle of predictive modeling with memory in an agent. For example, the gradient block between the policy and the MBP is not necessary (Ext.~Fig.~7); we have only included it to demonstrate the independence of these modules from each other. Similarly, it is possible to attach the policy as an extra feedforward network that receives $h_t$, $z_t$, and $m_t$ to the MBP. This worked comparably in terms of performance (Ext.~Fig.~8) but obscures how the system works conceptually since the independence of the MBP and policy is no longer explicit. On the other hand, lesions to the architecture, in particular removing the memory, the sensory or return prediction, or the retroactive memory updating, were harmful to performance or data efficiency (Ext.~Fig.~5).

Across all tasks, MERLIN performed qualitatively better than comparison models, which often entirely failed to store or recall information about the environment (Figs.~\ref{fig:memory_and_nav}a\&c; \ref{fig:battery}e-h; Ext.~Figs.~2\&9). On a few tasks, by increasing $\tau$ to ten times the duration used by MERLIN (for MERLIN $\tau=20$: $1.3$s at 15 frames per second), we were able to improve the performance of RL-MEM. However, this performance was always less than MERLIN's and was achieved after much more training (Ext.~Figs.~10-11). For the episodic Water Mazes task, extending $\tau$ was however unable to improve performance at all (Ext.~Fig.~10), just as for the Memory Game where $\tau$ was the duration of the whole episode. Memory systems trained end-to-end for performance (Eq.~\ref{eq:policy_gradient}) were less able to learn to write essential information into memory. 

This problem with end-to-end learning will only become more pressing as AI approaches the frontier of long-lived agents with long-term memories. For example, consider an agent that needs to remember events that occurred 24 hours ago. Stored memories could be retrieved by a memory reading operation, but methods for optimising network dynamics or information storage over that interval, like backpropagation-through-time, require keeping an exact record of network states over 24 hours. This is a stipulation that is practically prohibitive, and its neural implausibility suggests, at the very least, that there are better algorithmic solutions than end-to-end gradient computation for memory and other aspects of temporal credit assignment. We note that MERLIN exclusively used a window of $1.3$s to solve tasks requiring the use of memory over much longer intervals (i.e., 0.36\% of the length of the longest task at 6 minutes). 

While end-to-end RL with a sufficiently large network and enough experience and optimisation should theoretically learn to store relevant information in memory to call upon for later decisions, we have shown the actual requirements are prohibitive; as we have long known in domains such as object recognition and vision, architectural innovations (e.g. convolutional neural networks\cite{lecun1990handwritten}) are critical for practical systems. Although the implementation details will likely change, we believe that the combined use of memory and predictive modeling will prove essential to future large-scale agent models in AI and even in neuroscience\cite{eliasmith2012large}.

\bibliography{sample}
\bibliographystyle{Science}

\section*{Video Links}
\begin{enumerate}
    \item Ext.~Video~1: \url{https://youtu.be/YFx-D4eEs5A}
    \item Ext.~Video~2: \url{https://youtu.be/IiR_NOomcpk}
    \item Ext.~Video~3: \url{https://youtu.be/dQMKJtLScmk}
    \item Ext.~Video~4: \url{https://youtu.be/xrYDlTXyC6Q}
    \item Ext.~Video~5: \url{https://youtu.be/04H28-qA3f8}
    \item Ext.~Video~6: \url{https://youtu.be/3iA19h0Vvq0}
\end{enumerate}

\section*{Acknowledgments}
We thank David Silver, Larry Abbott, and Charles Blundell for helpful comments on the manuscript; Daan Wierstra, Neil Rabinowitz, Ari Morcos, Nicolas Heess, Alex Graves, Dharshan Kumaran, Raia Hadsell, Brian Zhang, Oriol Vinyals, and Hubert Soyer for discussions; Amir Sadik and Sarah York for environment testing; Stephen Gaffney and Helen King for organisational help. 
\section*{Author Contributions}
G.W. and T.L. conceived the project. G.W., C.C.H., D.A., M.M., A.A., A.S., M.G., S.M., D.R., and T.L. developed the model. C.C.H., D.A., G.W., M.M., A.A., and J.A. performed experiments. D.A., J.R., M.R., T.H., and D.Saxton built engineering infrastructure. C.C.H., G.W., D.A., M.M., A.A., A.G.B., and P.M. performed analyses. G.W., C.C.H., A.A., J.Z.L., and T.L. designed tasks. A.C. prepared figure artwork. K.K., M.B., D.Silver, and D.H. consulted for project. C.H. helped with team coordination. G.W., C.C.H., D.A., M.M., A.A., A.G.B., P.M., and T.L. wrote the manuscript. 
\section*{Correspondence}
Correspondence should be addressed to Greg Wayne, Chia-Chun Hung, or Timothy Lillicrap (email: $\{$gregwayne, aldenhung, countzero$\}$@google.com).

\newpage

\begin{center}
\textbf{\Large{Methods}}
\end{center}

\singlespacing
\tableofcontents

\newpage

\section{Environment Software}
All 3D environment experiments were implemented in DeepMind Lab (or DM Lab)~\cite{beattie2016deepmind}. For all experiments in this framework, we used frame rates of 60 frames per second but propagated only the first observation of each sequence of four to the networks. Rewards accumulated over each packet of four frames were summed together and associated with the first, undropped frame. Similarly, the agent chose one action at the beginning of this packet of four frames, and the action was repeated for all four steps. Accordingly, the number of ``Environment Steps'' reported in this manuscript is $4$ times the number of ``Agent Steps''.

We used a consistent action set for all experiments except for the Arbitrary Visuomotor Mapping task. For all other tasks, we used a set of six actions: \emph{move forward}, \emph{move backward}, \emph{rotate left with rotation rate of 30} (mapping to an angular acceleration parameter in DM Lab), \emph{rotate right with rotation rate of 30}, \emph{move forward and turn left}, \emph{move forward and turn right}. For the Arbitrary Visuomotor Mapping, we did not need to move relative to the screen, but we instead needed to move the viewing angle of the agent. We thus used four actions: \emph{look up}, \emph{look down}, \emph{look left}, \emph{look right} (with rotation rate parameter of 10).

\section{Model}
At a high level, the model consists of the memory-based predictor and the policy. The memory-based predictor contains within it several encoders and decoders and represents two distributions over the state variable: the prior, which predicts the distribution of the next state variable, and the posterior, which represents the distribution after the next observation. The memory-based predictor contains an external memory that stores a history of state variables. The policy takes the state variable as input and also reads from the external memory.

We now describe the model in detail by defining its parts list and the loss functions used to optimise it. Parameters given per task are defined in Table~\ref{table:merlin_parameters}.

\subsection{Encoder}
The encoder is composed of five sub-networks: the image encoder, the velocity encoder (in all DM Lab experiments), the action encoder, the reward encoder, and the text encoder. These act independently on the different elements contained within the input set $o_t \equiv (I_t, v_t, a_{t-1}, r_{t-1}, T_t)$, and their outputs are concatenated into the flat vector $e_t$. 

\subsubsection{Image Encoder}
For the DM Lab tasks, we use an image encoder that takes in image tensors of size $64 \times 64 \times 3$ (3 channel RGB). We then apply 6 ResNet~\cite{he2016deep} blocks with rectified linear activation functions. All blocks have 64 output channels and bottleneck channel sizes of 32. The strides for the 6 blocks are $(2, 1, 2, 1, 2, 1)$, resulting in 8-fold spatial down-sampling of the original image. Therefore, the ResNet module outputs tensors of size $8 \times 8 \times 64$. We do not use batch-norm~\cite{ioffe2015batch}, a pre-activation function on inputs, or a final activation function on the outputs. Finally, the output of the ResNet is flattened (into a $4,096$ element vector) and then propagated through one final linear layer that reduces the size to 500 dimensions, whereupon a $\tanh$ nonlinearity is applied.

For the task ``Memory'', we use the same architecture as for the DM Lab task save that the input image tensor size is $32 \times 32 \times 1$ (grey-scale). This produces ResNet module output tensors of size $4 \times  4 \times 64$. 

\subsubsection{Velocity Encoder}
The velocity encoder takes a 6-dimensional vector $v_t$ that comprises the translational and rotational velocities of the agent. It is calculated from measured differences between the current time step and the previous step (making it egocentric) but is zero across any agent teleportation. Though there is no physical distance metric in the DM Lab environment, the actual numerical velocities produced by the simulator are large, so we scale all numbers by dividing by 1000 before passing the data through the encoder. The encoder then applies a linear layer with an output dimensionality of 10. 

\subsubsection{Action Encoder}
In all environments, the action from the previous time step is a one-hot binary vector $a_{t-1}$ (6-dimensional for most DM Lab tasks) with $a_0 \equiv 0$. We use an identity encoder that leaves the action one-hot unchanged.

\subsubsection{Reward Encoder}
The reward from the previous time step $r_{t-1}$ is represented as a scalar magnitude and is not processed further.

\subsubsection{Text Encoder}
To interpret text instructions, a small, single layer LSTM that processes the word tokens (vocabulary size 1000) in the input by first passing through a linear embedding of size 50 that is then input into an LSTM of width 100. The terminal hidden state is treated as the output. 

\subsection{Decoder}
The decoder is composed of six sub-networks.
Five of these sub-networks are duals (every layer has dimensions transposed from the corresponding encoder layer) of the encoder networks.
The additional sub-network decodes a prediction of the return.

\subsubsection{Image Decoder}
The image decoder has the same architecture as the encoder except the operations are reversed. In particular, all 2D convolutional layers are replaced with transposed convolutions~\cite{dumoulin2016guide}. Additionally, the last layer produces a number of output channels that is formatted to the likelihood function used for the image reconstruction loss, described in more detail in Eq.~\ref{eq:output_loss}.

\subsubsection{Return Prediction Decoder and Value Function}
The return prediction decoder is the most complicated non-visual decoder. It is composed of two networks. The first is a value function MLP that takes in the concatenation of the latent variable with the policy distribution's multinomial logits $[z_t, \log \pi_\theta(a_t | M_{t-1}, z_{\leq t})]$. This vector is propagated through a single hidden layer of $200$ hidden units and a $\tanh$ activation function, which then 
projects via another linear layer to a 1-dimensional output. This function can be considered a state-value function
$V_t^\pi \equiv V^\pi(z_t, \log \pi_t)$.

A second network acts as a state-action advantage function, taking in the concatenation $[z_t, a_t]$ and propagating through an MLP with two hidden layers each of size 50 and $\tanh$ nonlinearities: $A^\pi(z_t, a_t)$. This quantity is then added to the value function $V^\pi$ to produce a return prediction $\hat{R}^\pi_t = \sg{V^\pi_t} + A^\pi_t$. When we calculate loss gradients, we do not allow gradients to flow back through the value function $V^\pi_t$ from the return prediction cost, as this changes the effective weight of the gradient on the value function term, which has its own loss function. The return prediction $\hat{R}^\pi_t$ can also be considered a state-action value function for the current policy (a Q-function).

\subsubsection{Text Decoder}
We decode the multi-word instruction sequentially with a single layer LSTM of the same size as the corresponding encoder whose input is $z_t$ and whose output is a 1,000-way softmax over each word token in the sequence. 

\subsubsection{Other Decoders}
The reward, velocity, and action decoders are all linear layers from $z_t$ to, respectively, $1$ dimension, $6$ dimensions, and the action cardinality.

\subsection{Memory}
Our memory system is primarily based on a simplification of the Differentiable Neural Computer (DNC)~\cite{graves2016hybrid}.
The memory itself is a two-dimensional matrix $M_t$ of size $(\nm, 2 \times |z|)$, where $|z|$ is the dimensionality of the latent state vector. The memory at the beginning of each episode is initialised blank, namely $M_0 = 0$. 

\subsubsection{Deep LSTMs}
In both the policy and memory-based predictor, we use a deep LSTM~\cite{graves2013speech} of two hidden layers. Although the deep LSTM model has been described before, we describe it here for completeness. Denote the input to the network at time step $t$ as $x_t$. Within a layer $l$, there is a recurrent state $h^l_t$ and a ``cell'' state $s_t^l$, which are updated based on the following recursion (with $\sigma(x) \equiv (1 + \exp(-x))^{-1}$):
\begin{align*}
i^l_t &= \sigma\left(W_i^l [x_t, h^l_{t-1},h^{l-1}_{t}]  + b^l_i \right)\\
f^l_t &= \sigma\left(W_f^l [x_t, h^l_{t-1},h^{l-1}_{t}]  + b^l_f \right)\\
s^l_t &= f_t^l s_{t-1}^l + i_t^l \tanh \left(W_s^l [x_t, h^l_{t-1},h^{l-1}_{t}] + b^l_s  \right)\\
o^l_t &= \sigma\left(W_o^l [x_t, h^l_{t-1},h^{l-1}_{t}]  + b^l_o \right)\\
h^l_t &= o^l_t \tanh(s^l_t) 
\end{align*}

In both uses of the deep LSTM (policy and MBP), to produce a complete output $h_t$, we concatenate the output vectors from each layer: $h_t \equiv [h_t^1, h_t^2]$. These are passed out for downstream processing.

\subsubsection{MBP LSTM}
At each time step $t$, the recurrent network of the MBP receives concatenated input
$[z_t, a_t]$, where $a_t$ is represented by a one-hot code. The policy network receives input $z_t$. For the MBP LSTM, the input is concatenated with a list of $\nkey$ vectors read from the memory at the previous time step $m_{t-1} \equiv [m_{t-1}^1, m_{t-1}^2, \dots, m_{t-1}^{\nkey}]$. The input to the MBP LSTM is therefore $x_t = [z_t, a_t, m_{t-1}]$. The deep LSTM equations are applied, and the output $h_t = [h_t^1, h_t^2]$ is produced. A linear layer is applied to the output, and a memory interface vector $i_t$ is constructed of dimension $\nkey \times (2 \times |z| + 1)$, where $|z|$ is the dimensionality of the latent vector. $i_t$ is then segmented into $\nkey$ read key vectors $k_t^1, k_t^2, \dots, k_t^{\nkey}$ of length $2 \times |z|$ and $\nkey$ scalars ${sc}_t^1, \dots, {sc}_t^{\nkey}$, which are passed through the function $\text{SoftPlus}(x) = \log(1 + \exp(x))$ to create the scalars $\beta_t^1, \dots, \beta_t^{\nkey}$. 

\subsubsection{Reading}
Memory reading is executed before memory writing. Reading is content-based, and proceeds by first computing the cosine similarity between each read key and each memory row $j$: $c^{ij}_t = \cos(k_t^i, M_{t-1}[j, \cdot]) = \frac{k_t^i \cdot M_{t-1}[j, \cdot]}{|k_t^i| |M_{t-1}[j, \cdot]|}$. For each read key, a normalised weighting vector of length $\nm$ is then computed: 
\begin{align*}
w_t^i[j] & = \frac{\exp (\beta_t^i c^{ij}_t)}{\sum_{j'} \exp (\beta_t^i c^{ij'}_t)}
\end{align*}
For that key, the readout from memory is $m_t^i = M_{t-1}^\top w_t^i$. These readouts are concatenated together with the state of the deep LSTM $[h_t^1, h_t^2, m_t^1, \dots, m_t^{\nkey}]$ and output from the module.

\subsubsection{Writing}
After reading, writing to memory occurs, which we also define using weighting vectors. The write weighting $\ww_t$ has length $\nm$ and always appends information to the $t$-th row of the memory matrix at time $t$, i.e., $\ww_t[i] = \delta_{it}$ (using the Kronecker delta). A second weighting for retroactive memory updates forms a filter of the write weighting
\begin{align*}
\wre_t & = \gamma \wre_{t-1} + (1 - \gamma) \ww_{t-1},
\end{align*} where $\gamma$ is the same as the discount factor for the task.
Given these two weightings, the memory update can be written as an online update
\begin{align}
M_t & = M_{t-1} + \ww_t [z_t, 0]^\top + \wre_t [0, z_t]^\top,   
\label{eq:mem_update}
\end{align}
where $0$ is the zero-vector of length $|z|$. Each of these weightings is initialised so that $\ww_0 = \wre_0 = 0$.

In case the number of memory rows is less than the episode length, overwriting of rows is necessary. To implement this, each row $k$ contains a usage indicator: $u_t[k]$. This indicator is initialised to $0$ until the row is first written to. Subsequently, the row's usage is increased if the row is read from by any of the reading heads $u_{t+1}[k] = u_{t}[k] + \sum_i w_{t+1}^i[k]$. When allocating a new row for writing, the row with smallest usage is chosen.

\subsection{Prior Distribution}
The prior distribution is produced by an MLP that takes in the output from the MBP LSTM at the previous time step $[h_{t-1}, m_{t-1}]$ and passes it through two hidden layers with $\tanh$ activation functions of width $2 \times |z|$. There is a final linear layer that produces a diagonal Gaussian distribution for the current time step $(\mu_{t}^{\text{prior}}, \log \Sigma_{t}^{\text{prior}})$, where both the mean and log-standard deviation are of size $|z|$.

\subsection{Posterior Distribution}
The posterior distribution is produced in two stages. First, the outputs of the encoded features, the outputs of the MBP LSTM, and the prior distribution parameters are concatenated into one large vector 
\begin{align*}
n_t & = [e_t, h_{t-1}, m_{t-1}, \mu_{t}^{\text{prior}}, \log \Sigma_{t}^{\text{prior}}].
\end{align*}
This concatenated vector is then propagated through an MLP with two hidden layers of size $2 \times |z|$ and $\tanh$ activation functions, followed by a single linear layer that produces an output of size $2 \times |z|$. This MLP's function $f^\text{post}(n_t)$ is added to the prior distribution to determine the posterior distribution: $[\mu_t^\text{post}, \log \Sigma_t^\text{post}] = f^\text{post}(n_t) + [\mu_{t}^{\text{prior}}, \log \Sigma_{t}^{\text{prior}}]$.

\subsection{State Variable Generation}
After the posterior distribution is computed, the state variable $z_t$ is sampled as $z_t := \mu_t^\text{post} + \exp(\log \Sigma_t^\text{post}) \odot \xi_t$, where $\xi_t \sim \mathcal{N}(0,1)$ and `$\odot$' represents element-wise multiplication.

\subsection{Policy}
The operation of the policy is similar to that of the MBP LSTM. At time step $t$, before the MBP LSTM operates, the policy receives $z_t$. A deep LSTM that can also read from the memory in the same way as the MBP executes one cycle, but using only one read key, giving outputs $[\tilde{h}_t, \tilde{m}_t]$. These outputs are then concatenated again with the latent variable $[z_t, \tilde{h}_t, \tilde{m}_t]$ and passed through a single hidden-layer MLP with 200 $\tanh$ units. This then projects to the logits of a multinomial softmax with the dimensionality of the action space, which varies per environment (4-dimensional for the Arbitrary Visuomotor Mapping task, 6-dimensional for the rest of DM Lab, and 16-dimensional for Memory). The action $a_t$ is sampled and passed to the MBP LSTM as an additional input, as described above. 

\section{Derivation of the Variational Lower Bound}
The log marginal likelihood of a probabilistic generative model can be lower-bounded by an approximating posterior as a consequence of Jensen's inequality:
$\log p(x) \geq \mathbb{E}_{q(z|x)}\log  \bigg[\frac{p(x,z)}{q(z|x)}\bigg]$. For convenience, we define $z_{0:t} \equiv (z_0, z_1, z_2, \dots, z_t)$ and $z_{0:-1} \equiv \varnothing$ (the empty set).
Then for a temporal model that factorises as $p(x_{0:t}, z_{0:t}) = \prod_{\tau=0}^t p(x_\tau | z_\tau) p(z_\tau | z_{0:\tau-1})$ and approximate posterior $q(z_{0:t} | x_{0:t}) = \prod_{\tau=0}^t q(z_\tau | z_{0:\tau-1}, x_{0:\tau})$, this becomes
\begin{align*}
\log p(x_{0:t}) & \geq \mathbb{E}_{q(z_{0:t} | x_{0:t})} \log \frac{p(x_{0:t}, z_{0:t})}{q(z_{0:t} | x_{0:t})} \\
& = \mathbb{E}_{q(z_{0:t} | x_{0:t})} \log \frac{\prod_{\tau=0}^t p(x_\tau | z_\tau) p(z_\tau | z_{0:\tau-1})}{\prod_{\tau=0}^t q(z_\tau | z_{0:\tau-1}, x_{0:\tau})} \\
& = \mathbb{E}_{q(z_{0:t} | x_{0:t})} \sum_{\tau=0}^t\bigg[\log p(x_\tau | z_\tau) + \log p(z_\tau | z_{0:\tau-1}) - \log q(z_\tau | z_{0:\tau-1}, x_{0:\tau})\bigg] \\
& = \sum_{\tau=0}^t \mathbb{E}_{q(z_{0:t} | x_{0:t})} \bigg[\log p(x_\tau | z_\tau) + \log p(z_\tau | z_{0:\tau-1}) - \log q(z_\tau | z_{0:\tau-1}, x_{0:\tau})\bigg] \\
& = \sum_{\tau=0}^t \mathbb{E}_{q(z_{0:\tau} | x_{0:\tau})} \bigg[\log p(x_\tau | z_\tau) + \log p(z_\tau | z_{0:\tau-1}) - \log q(z_\tau | z_{0:\tau-1}, x_{0:\tau})\bigg] \\
& = \sum_{\tau=0}^t \mathbb{E}_{q(z_{0:\tau-1} | x_{0:\tau-1})} \mathbb{E}_{q(z_\tau | z_{0:\tau-1}, x_{0:\tau})} \bigg[\log p(x_\tau | z_\tau) + \log p(z_\tau | z_{0:\tau-1}) \\ & \hspace{6cm} - \log q(z_\tau | z_{0:\tau-1}, x_{0:\tau})\bigg] \\
& = \sum_{\tau=0}^t \mathbb{E}_{q(z_{0:\tau-1} | x_{0:\tau-1})} \bigg[\mathbb{E}_{q(z_\tau | z_{0:\tau-1}, x_{0:\tau})} \log p(x_\tau | z_\tau) \\ & \hspace{3.5cm} - \text{D}_\text{KL}[q(z_\tau | z_{0:\tau-1}, x_{0:\tau}) || p(z_\tau | z_{0:\tau-1})] \bigg].
\end{align*}

Suppose now that we partition the target variables into two sets $x$ and $y$. In the stationary case, we can still form the inequality
$\log p(x,y) \geq \mathbb{E}_{q(z|x)}\bigg[\frac{\log p(x,y,z)}{q(z|x)}\bigg]$, which does not condition the approximate posterior on one of the variables. Likewise, in the temporal case, we have 
\begin{align*}
\log p(x_{0:t}, y_{0:t}) & \geq
\sum_{\tau=0}^t \mathbb{E}_{q(z_{0:\tau-1} | x_{0:\tau-1})} \bigg[\mathbb{E}_{q(z_\tau | z_{0:\tau-1}, x_{0:\tau})} \log p(x_\tau, y_\tau | z_\tau) \nonumber \\ & \hspace{3.5cm} - \text{D}_\text{KL}[q(z_\tau | z_{0:\tau-1}, x_{0:\tau}) || p(z_\tau | z_{0:\tau-1})] \bigg].
\end{align*}
This is the form we use to justify a loss function that combines prediction of incrementally observable information (image, reward, etc.) with information that is only known with some delay (the sum of future rewards) and therefore cannot be conditioned on in a filtering system. Finally, we can additionally condition the prior model on other variables such as actions $p(z_\tau | z_{0:\tau-1}, a_{0:\tau-1})$, giving
\begin{align}
\log p(x_{0:t}, y_{0:t}) & \geq
\sum_{\tau=0}^t \mathbb{E}_{q(z_{0:\tau-1} | x_{0:\tau-1})} \bigg[\mathbb{E}_{q(z_\tau | z_{0:\tau-1}, x_{0:\tau})} \log p(x_\tau, y_\tau | z_\tau) \nonumber \\ & \hspace{1cm} - \text{D}_\text{KL}[q(z_\tau | z_{0:\tau-1}, x_{0:\tau}) || p(z_\tau | z_{0:\tau-1}, a_{0:\tau-1})] \bigg].
\label{eq:partial_vlb}
\end{align}

\section{Cost Functions for the MBP and Policy}
\label{sec:mbp_and_policy_loss}
The parameters of the policy are entirely independent of the parameters of the memory-based predictor; they are not updated based on gradients from the same loss functions. This is implemented via a gradient stop between the policy and the state variable $z_t$. 

The memory-based predictor has a loss function based on the variational lower bound in Eq.~\ref{eq:partial_vlb} with specific architectural choices for the output model (the decoders alongside the likelihood functions for each prediction) and prior and posterior distributions:
\begin{align}
\log p(o_{0:t}, R_{0:t}) & \geq \sum_{\tau=0}^t \mathbb{E}_{q(z_{0:\tau-1} | o_{0:\tau-1})} \bigg[\mathbb{E}_{q(z_{\tau} | z_{0:\tau-1}, o_{0:\tau})} \log p(o_\tau, R_\tau | z_\tau) \nonumber \\ & \hspace{1cm}- \text{D}_\text{KL}[q(z_\tau | z_{0:\tau-1}, o_{0:\tau}) || p(z_\tau | z_{0:\tau-1}, a_{0:\tau-1})] \bigg]. 
\label{eq:return_vlb}
\end{align}

\subsection{Conditional Log-Likelihood}
The conditional log-likelihood term $\log p(o_t, R_t | z_t)$ is factorised into independent loss terms associated with each decoder and is conditioned on a sample $z_t$ from the approximate posterior network, thus giving a stochastic approximation to the expectation in the variational lower bound objective. We use a multinomial softmax cross-entropy loss for the action, mean-squared error (Gaussian with fixed variance of 1) losses for the velocity (if relevant), reward, and return information, a Bernoulli cross-entropy loss for each pixel channel of the image, and a multinomial cross-entropy for each word token. Thus, we have a negative conditional log-likelihood loss contribution at each time step of
\begin{align}
- \log p(o_t, R_t | z_t) & \equiv \alpha_\text{image} \mathcal{L}_\text{image}
+ \alpha_\text{return} \mathcal{L}_\text{return} + \alpha_\text{reward} \mathcal{L}_\text{reward} \nonumber \\
& + \alpha_\text{action} \mathcal{L}_\text{action} + \alpha_\text{velocity} \mathcal{L}_\text{velocity} + \alpha_\text{text} \mathcal{L}_\text{text},
\label{eq:output_loss}
\end{align}
where 
\begin{align}
\mathcal{L}_\text{image} & = \sum_{w=1,h=1,c=1}^{|W|,|H|,|C|} \bigg[ I_t[w,h,c] \log \hat{I}_t[w,h,c] + (1 - I_t[w,h,c]) \log (1 - \hat{I}_t[w,h,c] ) \bigg] \nonumber, \\
\mathcal{L}_\text{return} & = \frac{1}{2} \bigg [ || R_t - V^\pi(z_t, \log \pi_t) ||^2 + ||R_t - (\sg{V^\pi(z_t, \log \pi_t)} + A^\pi(z_t, a_t)) ||^2 \bigg ], \nonumber \\
 \mathcal{L}_\text{reward} & = \frac{1}{2} || r_{t-1} - \hat{r}_{t-1} ||^2, \nonumber \\
\mathcal{L}_\text{action} & = \sum_{i=1}^{|A|} \bigg [a_{t-1}[i] \log ( \hat{a}_{t-1}[i]) + (1-a_{t-1}[i]) \log (1 - \hat{a}_{t-1}[i]) \bigg ] \nonumber, \\
 \mathcal{L}_\text{velocity} & = \frac{1}{2}\sum_{i=1}^6 || v_t[i] - \hat{v}_t[i] ||^2 \nonumber, \\
\mathcal{L}_\text{text} & = \sum_{k=1}^{10} \sum_{i=1}^{1,000} \bigg [T_{t}[k,i] \log \hat{T}_t[k,i] + (1-T_{t}[k,i]) \log (1 - \hat{T}_t[k,i]) \bigg ] \nonumber.
\end{align}
In the text loss, $k$ indexes the word in the sequence (up to $10$ words in a string), and $1,000$ is the vocabulary size.
Constructing the target return value $R_t$ requires some subtlety. For environments with long episodes of length $T$, we use ``truncation windows''~\cite{mnih2016asynchronous} in which the time axis is subdivided into segments of length $\tau_\text{window}$. If the window around time index $t$ ends at time index $k$, the return within the window is 
\begin{align}
R_t =
\begin{cases}
r_t + \gamma r_{t+1} + \gamma^2 r_{t+2} + \dots + \gamma^{k - t + 1} V^\pi_\nu(z_{k + 1}, \log \pi_{k + 1}), \text{ if } k < T,\\
r_t + \gamma r_{t+1} + \gamma^2 r_{t+2} + \dots + \gamma^{T - t} r_T, \text{ if } T \leq k.
\end{cases}
\label{eq:bootstrap}
\end{align}

\subsection{Kullback-Leibler Divergence} The Kullback-Leibler Divergence term in Eq.~\ref{eq:return_vlb} is computed as the analytical KL Divergence between the two diagonal Gaussian distributions specified by the posterior and the prior networks:
\begin{align*}
\text{D}_\text{KL}[q(z_\tau | z_{0:\tau-1}, o_{0:\tau}) || p(z_\tau | z_{0:\tau-1}, a_{0:\tau-1})] & \equiv \text{D}_\text{KL}[\mathcal{N}(\mu_{t}^{\text{q}}, \log \Sigma_{t}^{\text{q}}) || \mathcal{N}(\mu_{t}^{\text{p}}, \log \Sigma_{t}^{\text{p}})].
\end{align*}

\subsection{Practical Details}
The contribution of each time step to the loss function in Eq.~\ref{eq:return_vlb} is
\begin{align}
\mathcal{L}_t & \geq \mathbb{E}_{z_\tau \sim \mathcal{N}(\mu_{t}^{\text{q}}, \log \Sigma_{t}^{\text{q}})} [\log p(o_\tau, R_\tau | z_\tau)] - \text{D}_\text{KL}[\mathcal{N}(\mu_{t}^{\text{q}}, \log \Sigma_{t}^{\text{q}}) || \mathcal{N}(\mu_{t}^{\text{p}}, \log \Sigma_{t}^{\text{p}})].     
\label{eq:per_step}
\end{align}
As a measure to reduce the magnitude of the gradients, the total loss that is applied to the memory-based predictor is divided by the number of pixel-channels $|W| \times |H| \times |C|$.

\subsection{Policy Gradient}
The policy gradient computation is slightly different from the one described in the main text (Eq.~1). Instead, we use discount and bootstrapping parameters $\gamma$ and $\lambda$, respectively, as part of the policy advantage calculation given by the Generalised Advantage Estimation algorithm \cite{schulman2015high}. Defining $\delta_t \equiv r_t + \gamma V^\pi(z_{t+1}, \log \pi_{t+1}) - V^\pi(z_{t}, \log \pi_{t})$, Generalised Advantage Estimation makes an update of the form:
\begin{equation}
\Delta \theta \propto \sum_{t=k \tau}^{(k+1)\tau} \sum_{t'=t}^{(k+1) \tau} (\gamma \lambda)^{t'-t} \delta_{t'} \nabla_\theta \log \pi_\theta(a_t | h_t). \label{eq:policy_gradient_detailed}
\end{equation}
There is an additional loss term that increases the entropy of the policy's action distribution. This and pseudocode for all of MERLIN's updates are provided in Alg.~\ref{alg:merlin}.

\section{Comparison Models}
\subsection{RL-LSTM} This model shares the same encoder networks as MERLIN, acting on its input to produce the same vector $e_t$. This is then passed as input to a deep LSTM that is the same as the deep policy LSTM in MERLIN. The deep policy LSTM has two output ``heads'', which are linear outputs from the LSTM state $h$, as in A3C~\cite{mnih2016asynchronous}: one for the value function baseline (return prediction) and one for the action distribution. Unlike the optimisation prescription of A3C, the policy head is trained using Eq.~\ref{eq:policy_gradient_detailed}, and the value head is trained by return prediction.

\subsection{RL-DNC} This is the same as the RL-LSTM except that the deep LSTM is replaced by a Differentiable Neural Computer~\cite{graves2016hybrid}. This model has a component with quadratic computational complexity as the memory size is scaled, so it is only applied to the memory game where the episodes are short.

\subsection{RL-MEM} In this model, the deep LSTM of the policy outputs an additional write vector with the same size as the state variable in the corresponding MERLIN experiment. This is directly stored in the memory, just as the state variable for MERLIN is. There is no retroactive memory updating. Reading from memory by the policy works the same way memory reading is implemented in MERLIN's policy. 

\section{Implementation and Optimisation}
For optimisation, we used truncated backpropagation through time~\cite{sutskever2013training}. We ran 192 parallel worker threads that each ran an episode on an environment and calculated gradients for learning. Each gradient was calculated after one truncation window $\tau_\text{window}$, which for DM Lab experiments was less than the duration of an episode. For reinforcement learning, after every truncation window, we also ``bootstrapped'' the return value, as described in Eq.~\ref{eq:bootstrap}. The gradient computed by each worker was sent to a ``parameter server'' that asynchronously ran an optimisation step with each incoming gradient. The memory-based predictor and policy were optimised using two separate ADAM optimisers~\cite{kingma2014adam} with independent learning rates $\eta^\text{mbp}$ and $\eta^{\pi}$. 

The pseudocode for each MERLIN worker is presented in Alg.~\ref{alg:merlin}. For all experiments, we used the open source package Sonnet (\href{https://github.com/deepmind/sonnet}{https://github.com/deepmind/sonnet}) along with its defaults for parameter initialisation.

\begin{algorithm}
\caption{MERLIN Worker Pseudocode}
\label{alg:merlin}
\begin{algorithmic}
\State // Assume global shared parameter vectors $\theta$ for the policy network and $\chi$ for the memory-based predictor; global shared counter $T := 0$
\State // Assume thread-specific parameter vectors $\theta', \chi'$
\State // Assume discount factor $\gamma \in (0,1]$ and bootstrapping parameter $\lambda \in [0,1]$
\State Initialize thread step counter $t := 1$
\Repeat
\State Synchronize thread-specific parameters $\theta' := \theta; \chi' := \chi$
\State Zero model's memory \& recurrent state if new episode begins
\State $t_\text{start} := t$
\Repeat
\State Prior $\mathcal{N}(\mu_{t}^{\text{p}}, \log \Sigma_{t}^{\text{p}}) = p(h_{t-1}, m_{t-1})$
\State $e_t = \text{enc}(o_t)$
\State Posterior $\mathcal{N}(\mu_{t}^{\text{q}}, \log \Sigma_{t}^{\text{q}}) = q(e_t, h_{t-1}, m_{t-1}, \mu_{t}^{\text{p}}, \log \Sigma_{t}^{\text{p}})$
\State Sample $z_t \sim \mathcal{N}(\mu_{t}^{\text{q}}, \log \Sigma_{t}^{\text{q}})$ 
\State Policy network update $\tilde{h}_t = \text{rec}(\tilde{h}_{t-1}, \tilde{m}_{t}, \sg{z_t})$
\State Policy distribution $\pi_t = \pi(\tilde{h}_t, \sg{z_t})$
\State Sample $a_t \sim \pi_t$
\State $h_t = \text{rec}(h_{t-1}, m_t, z_t)$
\State Update memory with $z_t$ by Methods Eq.~\ref{eq:mem_update}
\State $R_t, o_t^{r} = \text{dec}(z_t, \pi_{t}, a_t)$
\State Apply $a_t$ to environment and receive reward $r_t$ and observation $o_{t+1}$
\State $t := t + 1; T := T + 1$
\Until{environment termination or $t-t_\text{start} == \tau_\text{window}$}
\State If not terminated, run additional step to compute
$V^\pi_\nu(z_{t+1}, \log \pi_{t+1})$ \State and set $R_{t+1} := V^\pi(z_{t+1}, \log \pi_{t+1})$ // (but don't increment counters)
\State Reset performance accumulators $\mathcal{A} := 0; \mathcal{L} := 0; \mathcal{H} := 0$
     \For{$k$ from $t$ down to $t_\text{start}$}
        \State $\gamma_t :=
            \begin{cases}
            0, \text{ if } k \text{ is environment termination} \\
            \gamma, \text{ otherwise } \\
            \end{cases}$ \\
        \State $R_k := r_k + \gamma_t R_{k+1}$      
        \State $\delta_k := r_k + \gamma_t V^\pi(z_{k+1}, \log \pi_{k+1}) - V^\pi(z_{k}, \log \pi_{k})$ 
        \State $A_k := \delta_k + (\gamma \lambda) A_{k+1}$
        \State $\mathcal{L} := \mathcal{L} + \mathcal{L}_k$ (Eq.~\ref{eq:per_step})
        \State $\mathcal{A} := \mathcal{A} + A_k \log \pi_k[a_k]$
        \State $\mathcal{H} := \mathcal{H} - \alpha_\text{entropy} \sum_i \pi_k[i] \log \pi_k[i]$ $\text{ (Entropy loss) }$
\EndFor
\State $d \chi' := \nabla_{\chi'} \mathcal{L}$
\State $d \theta' := \nabla_{\theta'} (\mathcal{A} + \mathcal{H})$
\State Asynchronously update via gradient ascent $\theta$ using $d \theta'$ and $\chi$ using $d \chi'$
\Until{$T > T_\text{max}$}
\end{algorithmic}
\end{algorithm}

\section{Tasks}
For all learning curves, each model was independently trained 5 times with different random seed initialisation. The learning curves are reported with standard errors across training runs, visualised as the shaded areas on the curves.

\subsection{Memory Game}
At each episode, eight pairs of cards are chosen from the Omniglot image dataset~\cite{lake2015human}. They are randomly placed on a $4 \times 4$ grid. The agent is given an initial blank observation, and at each turn the agent chooses a grid location for its action. If matching cards are selected on consecutive turns, a reward of 1 is given to the agent. In total, the agent is given 24 moves to clear the board, which is the maximum number of turns that would be needed by an optimal agent. To make the problem more challenging from the perspective of perception, we apply a random affine transformation to the image on the card each time it is viewed (a rotation uniformly between -0.2 and 0.2 radians, a translation on each axis of up to 2 pixels, and a magnification up to 1.15 times the size of the image). When the entire board is cleared, an additional bonus point is awarded at each timestep. If a previously cleared location is selected, the observation is shown as a blank. 

\subsection{Navigation Tasks}
These constitute a set of four tasks on which single agents are simultaneously trained. Each worker thread $w$ is assigned one of the four tasks to operate on using the formula $\text{task} = w \mod 4$.

\emph{Goal-Finding:} 
The Goal-Finding task tests if an agent can combine proximal and distal cues in order to find a goal in an environment with several rooms. Distal cues are provided by a fixed skyline of skyscrapers, and proximal cues are provided by wall and floor patterns and paintings on walls. At each episode, from 3 to 5 rooms are randomly placed down on cells in an arena of size $11$ units squared. (The agent moves with a much finer, nearly continuous scale.) Once the rooms are constructed, random corridors with distinct floor and wall patterning are built that connect some of the rooms. A single goal object is placed at random in one of the rooms. When the agent is first teleported in a random room, it has 90 seconds to find the goal, which is worth 10 points. On arriving at the goal, the agent will be teleported at random in one of the rooms, and must find the goal again. Returning to the goal faster on subsequent teleportations is evidence that the goal location has been memorised by the agent. 

\emph{No Skyline:}
The second task involves eliminating the distal cues provided by the skyline, demanding that the agent navigate only by means of the proximal cues. Otherwise, the task is the same as the Goal-Finding task.

\emph{Large Environment:}
In this task, the environment arena is procedurally generated on a $15$ units squared grid with up to 7 rooms.

\emph{Dynamic Doors:}
Entrances to rooms are barricaded by doors that are opened and closed randomly at every teleporation event. This task tests if agents are able to find robust navigation strategies involving replanning trajectories around obstacles.

\subsection{Arbitrary Visuomotor Mapping}
The task setting emulates a human or monkey psychophysics experiment. The agent views a screen and can tilt its view continuously in any of the four cardinal directions. On screen, stimuli from a human visual memory capacity experiment~\cite{brady2008visual} are shown to the agent in an experimental procedure known as an ``arbitrary visuomotor mapping'' experiment~\cite{wise2000arbitrary}. At first presentation of a stimulus, one of four targets in the cardinal directions lights up green, indicating the correct direction to move the agent's gaze. If the agent moves its gaze to the correct target, it is given a reward of $10$ points, and it receives $-1$ points for the wrong choice. When the agent is shown the same image subsequently, the targets are black before answering, and the agent must remember the appropriate choice direction. When it does again answer, the black targets briefly flash green and red to indicate the correct and incorrect targets respectively. Initially, there is only $k=1$ images in the set of possible query images for a trial, but when the agent answers $k$ times consecutively with the correct answer, a new image is added to the pool of images that must be remembered. A block of experimental trials runs continuously in a self-paced manner for 90 seconds.

\subsection{Rapid Reward Valuation}
The task takes place in an open arena of $11$ units squared in area.
Each trial lasts for 90 seconds and is composed of sub-episodes in which eight distinct objects (randomly assembled from a set of 16 base objects and a large number of textures) are randomly placed in the arena. These eight objects are randomly assigned values of $+2$ or $-1$ points, and we mandate that there be at least two positively and two negatively valued objects. Throughout the episode, the object value is fixed, but once all the positively valued objects have been collected, a new sub-episode begins with the same eight objects re-appear in permuted locations, allowing the agent to run up a high score. To solve this task, the agent must explore to identify good and bad objects, bind values with reasonably view-invariant representations of object identity, and use its memory to inform its decisions.

\subsection{Episodic Water Mazes}
This is a version of the Morris water maze experiment~\cite{morris1984developments} with the additional wrinkle that the agent must remember three mazes at a time, each distinguishable by wall colouring. In each water maze, an invisible platform is randomly placed within the arena, and the circular perimeter of the arena has fixed paintings to provide landmark cues. When the agent reaches the platform, the platform elevates automatically, and five consecutive rewards of 1 point are delivered for maintaining position on it. After this, the agent is teleported randomly in one of the three water mazes, continuing its activity for 360 seconds per episode.

\subsection{Executing Transient Instructions}
\label{sec:lang}
This task primarily demonstrates the MERLIN agent's versatility in a more abstract memory task that models the kind of instruction-following demonstrated by work dogs~\cite{pilley2011border}. It is a memory-dependent variant of an instruction-following task first presented by Hermann et al.~\cite{hermann2017grounded}. The map contains two rooms of different colours, separated by two corridors. Two coloured objects are placed in each room, and the agent is provided with an instruction in the format ``[object colour, room colour]'' for the first five time steps of the episode. The episode ends when the agent first collects an object, providing a reward of $+10$ for the correct object and $-1$ for the incorrect object. 

To interpret the instruction, the word tokens are sequentially processed by the LSTM encoder network. Its terminal hidden state is concatenated to the embedding vector $e_t$. Likewise, we decode the instruction sequentially with a single layer LSTM from $z_t$ (though this decodes $0$ after time step five). The loss coefficient for this model is $\alpha_\text{instruction}=0.5$.

\subsection{Latent Learning in a T-Maze}
This level is designed to demonstrate the model's capacity for latent learning. Each episode has three phases. In phase 1 (Fig.~5a left), the agent explores a T-Maze, where there are two random objects (randomly sampled from a set of 16 base objects and a large number of textures) at each arm of the T-maze. These two objects can not be consumed in phase 1; instead, the agent is required to collect four apples (worth $+1$ reward) to enter the next phase. In phase 2 (Fig.~5a middle), the agent is required to walk down a long corridor and collect apples along the route to enter the third phase. The point of this phase is to introduce an extended delay from the first (exploration) phase to the last (exploit) phase. In phase 3 (Fig.~5a right), the agent is teleported back into the original T-maze. There is a glass container in the middle of the maze presenting one of the two objects in phase 1. The agent can only go into one of the arms (as there are doors at the arms, and the agent must irreversibly select only one to open). The agent gets a $+10$ reward for collecting the object shown in the glass display and a $-3$ reward for collecting the other. The episode terminates when the agent collects one of the objects or 30 seconds has passed.

\subsection{Task Specific Parameters}
\label{sec:task_parameters}
For the LSTMs inside both RL comparisons and MERLIN, we used two layers of 256 units in DM Lab tasks and one layer of 50 units for Memory. In MERLIN, the memory-based predictor had three read keys, and the policy had 1 read key. In RL-MEM, the memory controller constructed 3 read keys as well. 

\begin{table}
\centering
 \begin{tabular}{||c c c c||} 
 \hline
 MERLIN & Memory & Arbit. Visuomotor Map. & Other DM Lab  \\ [0.5ex] 
 \hline\hline
$\eta^\text{mbp}$ & $10^{-5}$ & $5 \times 10^{-6}$ & $5 \times 10^{-6}$ \\ 
$\eta^\pi$  & $10^{-4}$ & $2 \times 10^{-5}$ & $2 \times 10^{-5}$ \\
$\gamma$ &1 & 0.8 & 0.96 \\
$\lambda$ & 0.8 & 0.9 & 0.9 \\ 
$\alpha_\text{velocity}$ & $\text{N/A}$ & 1 & 1 \\ 
$\alpha_\text{image}$ & 1 & 1 & 1 \\ 
$\alpha_\text{reward}$ & 1 & 1 & 1 \\ 
$\alpha_\text{return}$ & $1 \times \frac{1}{24}$ & $5 \times (1 - \gamma)$ & $5 \times (1 - \gamma)$ \\ 
$\alpha_\text{action}$ & 1 & 1 & 1 \\ 
$\alpha_\text{entropy}$ & 0.01 & 0.01 & 0.01 \\ 
$\tau_\text{window}$ & 24 & 20 & 20 \\ 
Size of $z$ & 100 & 200 & 200 \\ 
Retroact. Writing & N & Y & Y \\ 
$\nm$ & 40 & 1350 & 1350 \\ 
$\nkey_\text{mc}$ & 3 & 3 & 3 \\
$\nkey_\pi$ & 1 & 1 & 1 \\ [1ex] 
 \hline
 \end{tabular}
 \caption{\footnotesize{Task-specific parameters for MERLIN model.}}
 \label{table:merlin_parameters}
\end{table}

\begin{table}
\centering
 \begin{tabular}{||c c c c||} 
 \hline
 RL-\text{LSTM/MEM} & Memory & Arbit. Visuomotor Map. & Other DM Lab  \\ [0.5ex] 
 \hline\hline
$\eta$ & $10^{-5}$ & $5 \times 10^{-6}$ & $5 \times 10^{-6}$ \\ 
$\gamma$ &1 & 0.8 & 0.96 \\
$\lambda$ & 0.8 & 0.9 & 0.9 \\ 
$\alpha_\text{entropy}$ & 0.01 & 0.01 & 0.01 \\ 
$\tau_\text{window}$ & 24 & 20 & 20 \\ 
$\nm$ & 40 & 1350 & 1350 \\
Mem. Word Size & 100 & 200 & 200 \\[1ex] 
 \hline
 \end{tabular}
 \caption{\footnotesize{Task-specific parameters for RL models. The last two rows only apply to the RL-MEM model.}}
 \label{table:rl_parameters}
\end{table}

\section{Lesion Comparison Experiments}
In the lesion experiments, various parts of the model were removed as follows:
\begin{itemize}

\item `No memory': The external memory was removed. This left the prediction and policy parts of the model with separate deep LSTM recurrent components but no shared memory.

\item `Only return decoder': We removed the observation decoders and reconstruction costs; we also set the prior to $\mathcal{N}(0,1)$ and removed the Kullback-Leibler cost between the prior and posterior. The posterior network was retained, as was the return prediction decoder, so the model still had an objective to predict future returns.

\item `No return decoder': The return prediction decoder was removed from the generative part of the model. (A value function for policy gradient advantage calculation was learned as a separate linear projection from the policy.)

\item `No retroactive memory-update': The second column of the memory, which was updated with the discounted sum of future state variables, was removed.

\end{itemize}

\section{Task Analyses}

\subsection{Time-to-Goal Analysis}
For Fig.~3a, we analysed the time to goal in the Large Environment task for MERLIN and the two control models, RL-LSTM and RL-MEM. Time-to-goal was defined as the number of steps the agent took from been teleported to reaching the goal. We ran a common set of 200 test mazes for all three models and only included episodes where all three models reached the goal over 10 times. We then averaged the time-to-goal of teleported event $n$ for each model.

\subsection{Goal Location Decoding}
Coding of information related to both absolute (allocentric) and relative (egocentric) locations of the goal was analysed by decoding activity in each model~\cite{mirowski2016learning}. We trained the agent on the Large Environment version of the navigation task. We trained additional decoders that did not influence the representations of the model: they did not backpropagate error into the agent networks. 

Each agent was compared in two conditions: a base condition in which recurrent memory information was used for decoding; or a convnet condition in which the 500-dimensional output of the agent's image encoder was provided as input to the decoder. For the RL-LSTM base condition, the policy LSTM state $\tilde{h}_t$ was input to the decoder. For the RL-MEM base condition, the output of the policy LSTM and the reads of the policy LSTM $[\tilde{h}_t, \tilde{m}_t]$ were input to the decoder. For MERLIN, the input to the decoder was composed of several MBP variables $[z_t, h_t, m_t]$. 

Each decoder was a single hidden-layer perceptron with 256 hidden units. In the relative (egocentric) analysis, the decoder output was a multinomial classification over $16 \times 22$ expressing the goal in agent-relative polar coordinates: 16 elements for angle and 22 elements for distance-to-goal (Ext.~Fig.~1). To handle the wraparound of angles at $2 \pi$, the cosine between the angle represented by the bin and the measured angle to the goal was computed, then the arccosine of this quantity was used to produce the angular distance. In the absolute (allocentric) analysis, there were $15 \times 15$ bins for absolute $x$ position by absolute $y$ position (Fig.~3b).
In both figures, the prediction error was measured as the Earth-Mover distance between the ground truth location (delta function distribution over the correct bin) and the predicted distribution, using the Manhattan distance to measure the underlying metric distance. 

\subsection{Value Function Event-Triggered Analysis}
We applied this same methodology to produce Figs.~3d and 4k. The accuracy of the value function $V^\pi$ was evaluated by calculating the absolute difference between the agent's value function estimate and the empirical discounted return $|\delta_t| = |V^\pi_t-R_t|$. This calculation was registered to goal or object acquisition events, so that $|\delta(-t)|$ represents the absolute difference $t$ steps before reaching a goal. As above, we collected a dataset of 200 test episodes and plotted the average of $|\delta(-t)|$ for all goal-acquisitions that took at least 30 steps for the Large Environment navigation task or 15 steps for the Rapid Reward Valuation Task. 

\subsection{Variance Explained Analysis}
In Fig.~3f, we analyzed how much variance in the state variable $z$ could be explained by $\hat{V}^\pi$ for models trained with varied $\alpha_\text{return} \in (0.1, 1.0, 5.0, 10.0, 100.0, 1000.0)$. 
We collected 100 90-second episodes worth of data ($135,000$ data points) for each of the six models and trained a single hidden-layer perceptron with 200 hidden units to decode $z$ from $\hat{V}^\pi$, denoting the reconstruction $\hat{z}$. We then calculated the percentage variance explained: $\text{VE} = \frac{\text{Var}(z) - \text{MSE}(\hat{z})}{\text{Var}(z)}$. We then used 10-fold cross-validation to estimate $\text{VE}$ on the validation set and report the mean.

\subsection{Return Prediction Saliency}
To generate Fig.~3g (middle row), each pixel was shaded according to the L2 norm of the gradients across the three colour channels $\sum_{c=1}^3 |\partial V^\pi_t / \partial I^{w,h,c}_t|^2$, smoothed across the image with a Gaussian filter ($\sigma=2 \text{ pixels}$). The colour scale was chosen at the 0 and 99 percentiles of the gradient values across every pixel in the entire episode.

We also trained an agent that did not backpropagate return prediction error further than the state variable $z$. Thus, this model trained the decoder but could not modify the representations that were provided to it. After training, we removed the block on backpropagating return prediction error to analyse the model in the same way (Fig.~3g bottom row).

\subsection{Read Head Analysis in Navigation}
On the Large Environment task, for every time step $t$ of the simulation, we recorded the agent's 2D position in the maze and extracted the values of the read weightings used to read out from the memory. We mapped from each memory row to a spatial position in the environment $i \to (x,y)$ based on the spatial location of the agent at the time of writing to that row.

Across all agents we noticed that one read head focused around the current position of the agent; the two other heads read from positions closer to the goal location. 
We quantified this tendency in Fig.~3i by plotting the Euclidean distance of the read's centre-of-mass to the goal over the course of 50 steps before reaching the goal. Since some memory rows are not yet written at time $t$ (all locations with indices above $t$), we exclude these rows. 

We also indicate the distance of the agent to the goal in Fig.~3i, which shows that one of the heads always read from positions just a little closer to the goal than the agent. 

The statistics were collected from 200 test mazes, where the environment parameters were fixed so that each trained agent was exposed to the same sequence of mazes. We discarded the first 100 time steps of a run so that the agent had already visited the goal once or more. To produce Fig.~3i, we only used runs in which it took the agent at least 50 time steps to arrive at the goal, giving at least 1,200 sample runs per agent.  

\subsection{Rapid Reward Valuation Analyses}
In Fig.~4j, for a dataset of 50 episodes, we subselected all time points $t$ in which MERLIN was $k = 3$ time steps from consuming an object. Across all read heads, we identified the memory row $t_\text{past}$ with maximum weighting by any head. From $t_\text{past}$, which represents both a memory row index and a time step index, we searched forward for $g=15$ steps: i.e.,  $t_\text{past}+1, t_\text{past}+2, \dots, t_\text{past} + g$. We then calculated the percentage of cases in which either the same object was consumed at time $t+k$ as was consumed between $t_\text{past}$ and $t_\text{past}+g$ (same), a different object was consumed (different), or no object was consumed (null).

After the second consumption of an object, two values were collected: the state variable $z$ 3 time steps before the first consumption of the object and the retroactive memory $(1-\gamma) \sum_{t'>t} \gamma^{t'-t} z_{t'}$ written next to it. The data were collected after the second consumption of the object to ensure that enough time had elapsed for the retroactive memory update to be considered finalised. This process was repeated $25,000$ times. For either the state variable or the retroactive memory input, a logistic regression classifier was used to predict if the reward of the upcoming object was positive or negative, and the accuracy was estimated using five-fold cross-validation. The accuracy for the classifiers trained with state variable inputs was $63\%$. This was higher than $50\%$ because performant agents consumed good objects more than bad ones, biasing the input data. The accuracy for classifiers trained with retroactive memories as inputs was $93\%$.

\subsection{Episodic Water Mazes}
In Fig.~4l, we plotted a MERLIN agent's absolute position either on its first visit to the platform in each maze (red) or subsequent visits (green). In Ext.~Fig.~6, we also used 20 other episodes to perform a time-to-goal analysis as for the Large Environment navigation task. 

\subsection{Executing Transient Instructions}
In Fig.~4m, for every time step of the episode, we trained a separate decoder model that took as input the concatenation of the MBP recurrent state and reads $[h_{t},  m_{t}]$ ($1,712$ dimensions). Each decoder was a single hidden-layer MLP (hidden size $50$) whose output was required to predict a target of $91$ dimensions representing the composite instruction $<\text{room colour}> \times <\text{object colour}>$ (7 room colours by 13 room colours). We also trained two additional models to predict from the reads or MBP hidden state independently. From these models, we computed the average accuracy at predicting the instruction at each time step on a held-out set of $1,500$ episodes.

\subsection{Latent Learning in a T-Maze}
For each time step from the beginning of phase 3, we trained a separate logistic regression binary classifier to predict the goal location (left arm or right arm) from the MBP's memory reads $m_{t}$. Episodes in which the agent did not reach the final goal were removed. We reported the results of 5-fold cross-validation on a $2,000$ episode dataset.

\subsection{Spatial Receptive Fields of Memory Access}
In Ext.~Fig.~12, we calculated the spatial receptive fields of memory rows accessed by the third MBP read key in one episode. The environment (Large Environment) was divided into a $50 \times 50$ overhead grid. Each panel corresponds to a memory row $i$. For each $i$, we calculated the average value of the read weighting $w^i_t$ over the time points when the agent was at each grid location. We limited the data to memory rows written to after the first acquisition of the goal. Shown are the most active memory rows and the $10\%$ most active grid locations. The most active grid locations for each row tended to be close to the location of the initial write to that row.

\section{Human Testing}
For the navigation tasks, two professional game testers with prior practice on the standard environment variant of the task and knowledge of the structure of each task were given 15 minutes to practice again on each task. The testers then recorded 15 consecutive runs of 90 seconds each. The average of the scores of the human testers is reported.

For the Arbitrary Visuomotor Mapping task, the task creator (JZL) performed the task for 6 blocks of 90 seconds on both the naturalistic images and the synthetic transfer set. 

\break
\clearpage

\setcounter{figure}{0}    
\setcounter{table}{0} 
\renewcommand{\figurename}{Extended Figure}
\renewcommand{\tablename}{Extended Table}

\begin{center}
\textbf{\Large{Extended Figures}}
\end{center}

\begin{figure}[h!]
\center
\includegraphics[width=1\textwidth]{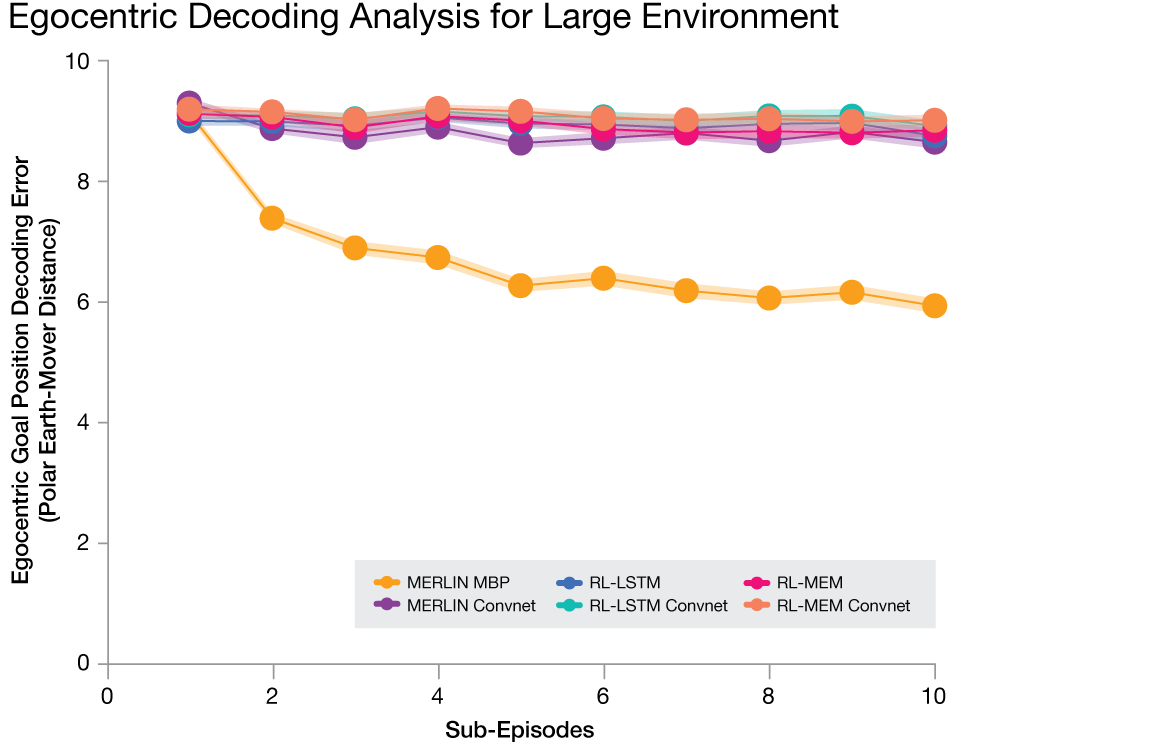}
\caption{\footnotesize{From MERLIN's MBP ($m_t$, $h_t$, and $z_t$), we could use logistic regression to classify the egocentric $(d,\theta)$ position of the goal to much better accuracy than could be achieved for the representations in the RL models or any of the convnet encoder output layers. See Methods Sec.~9.2.}}
\end{figure}

\begin{figure}[h!]
\center
\includegraphics[width=0.8\textwidth]{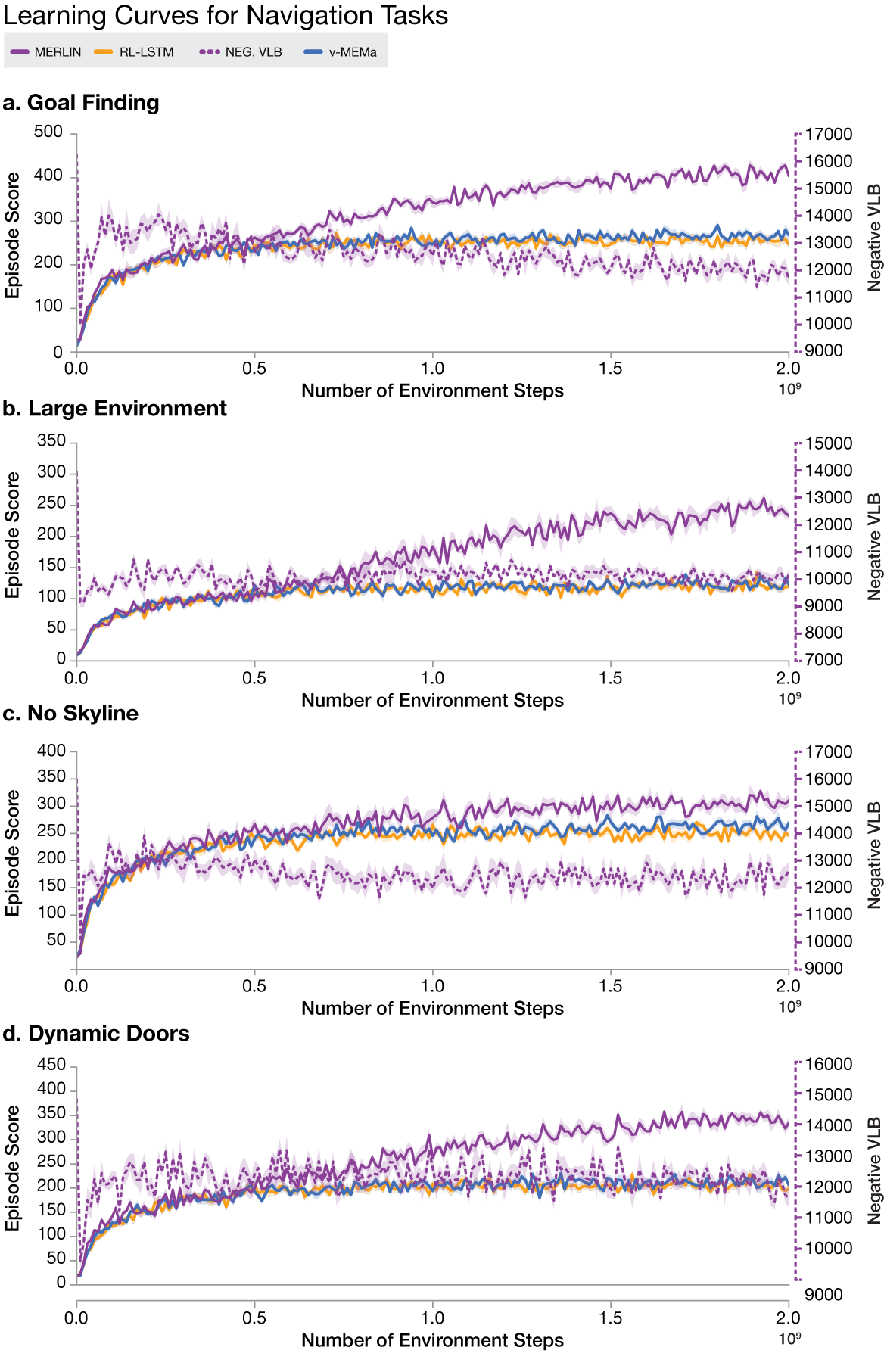}
\caption{\footnotesize{Learning Curves for Joint Training on All Navigation Tasks. Agents were trained on all of these tasks simultaneously. The number of environment steps was tallied across all workers on all tasks, so fewer steps were used per task than the time axis shows.}}
\end{figure}

\begin{figure}[h!]
\center
\includegraphics[width=1\textwidth]{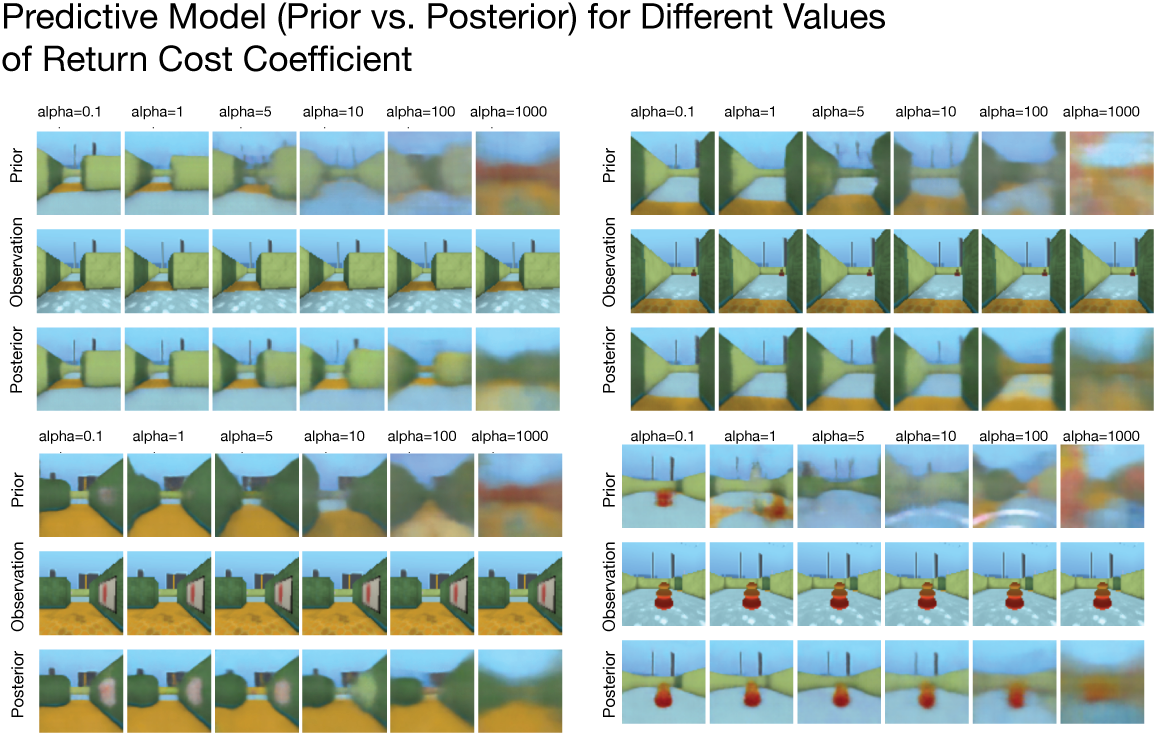}
\caption{\footnotesize{Four episodes in the Goal Finding environment are shown. In each we show the visual image predictions of the memory-based predictor for various values of the return prediction cost coefficient (alpha), from left to right ranging from $0.1$ to $1,000$. In each, the top row represents the mean of the predictive prior distribution $p$ for the state variable $\mu^p_{t+1}$, which is produced from information available at time $t$ but pertains to the visual image at time $t+1$. This mean is passed through the visual decoder to produce a visual image. In the middle row, we see the actual visual observation at time $t+1$. In the bottom row, we see the mean of the posterior distribution $q$ for the state variable $\mu^q_{t+1}$, which is produced given information available up to and including time $t+1$ (except for the return, which pertains to rewards occurring at $t+2$, $t+3$, ...). This is also passed through the visual decoder to produce a visual observation. In general, posterior distributions are sharper than prior distributions, as here, but the prior was a pretty good visual predictor (and match to the posterior) for low values of the return prediction cost coefficient ($r=0.1$, $r=1$). For larger values of the return cost coefficient, the MBP's prior predictions deteriorated; the agent performance however was not highest for the model that produced the best visual samples.}}
\end{figure}

\begin{figure}[h!]
\center
\includegraphics[width=0.75\textwidth]{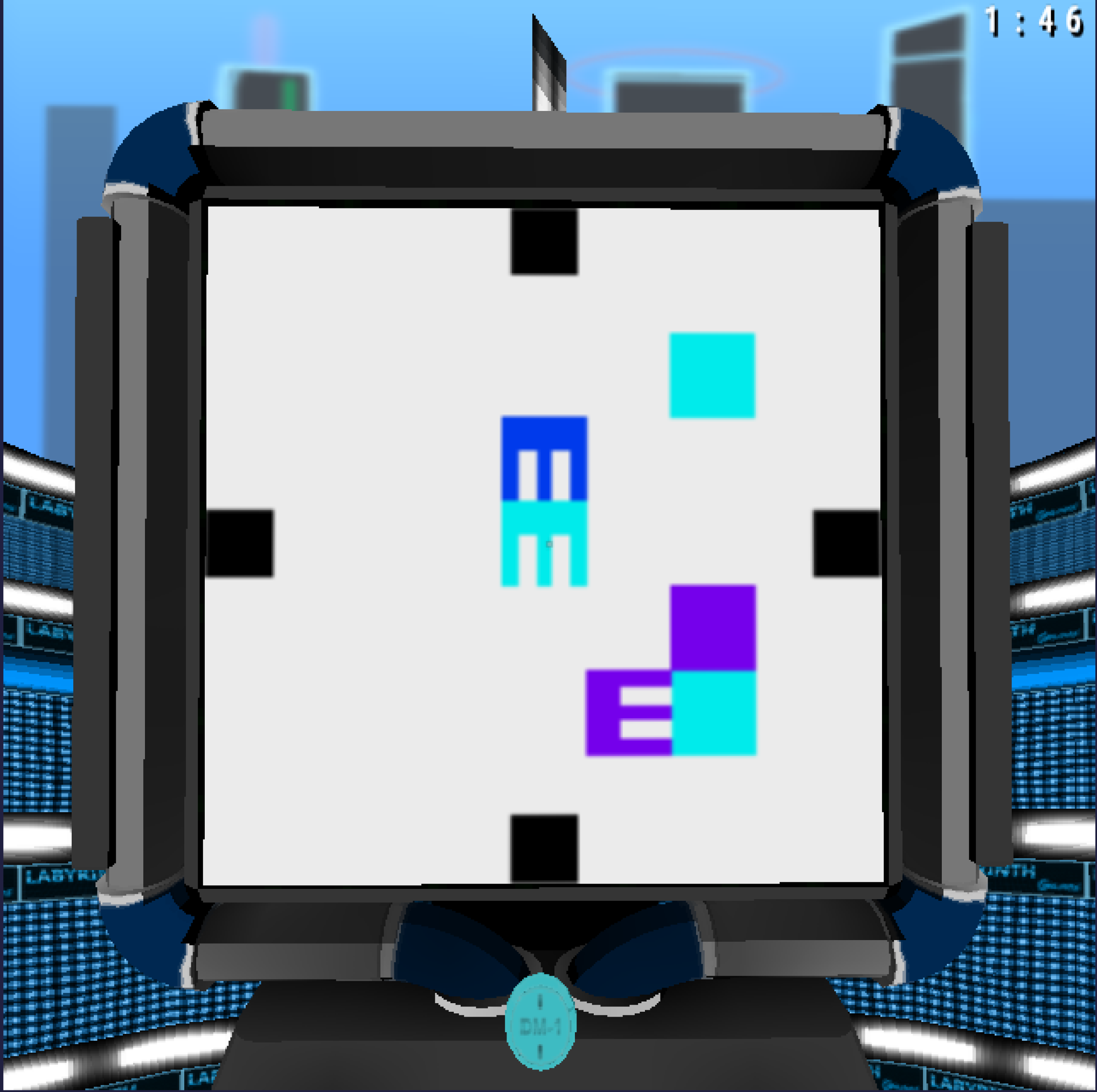}
\caption{\footnotesize{Arbitrary Visuomotor Mapping synthetic stimulus example used in transfer task.}}
\end{figure}

\begin{figure}[h!]
\center
\includegraphics[width=0.9\textwidth]{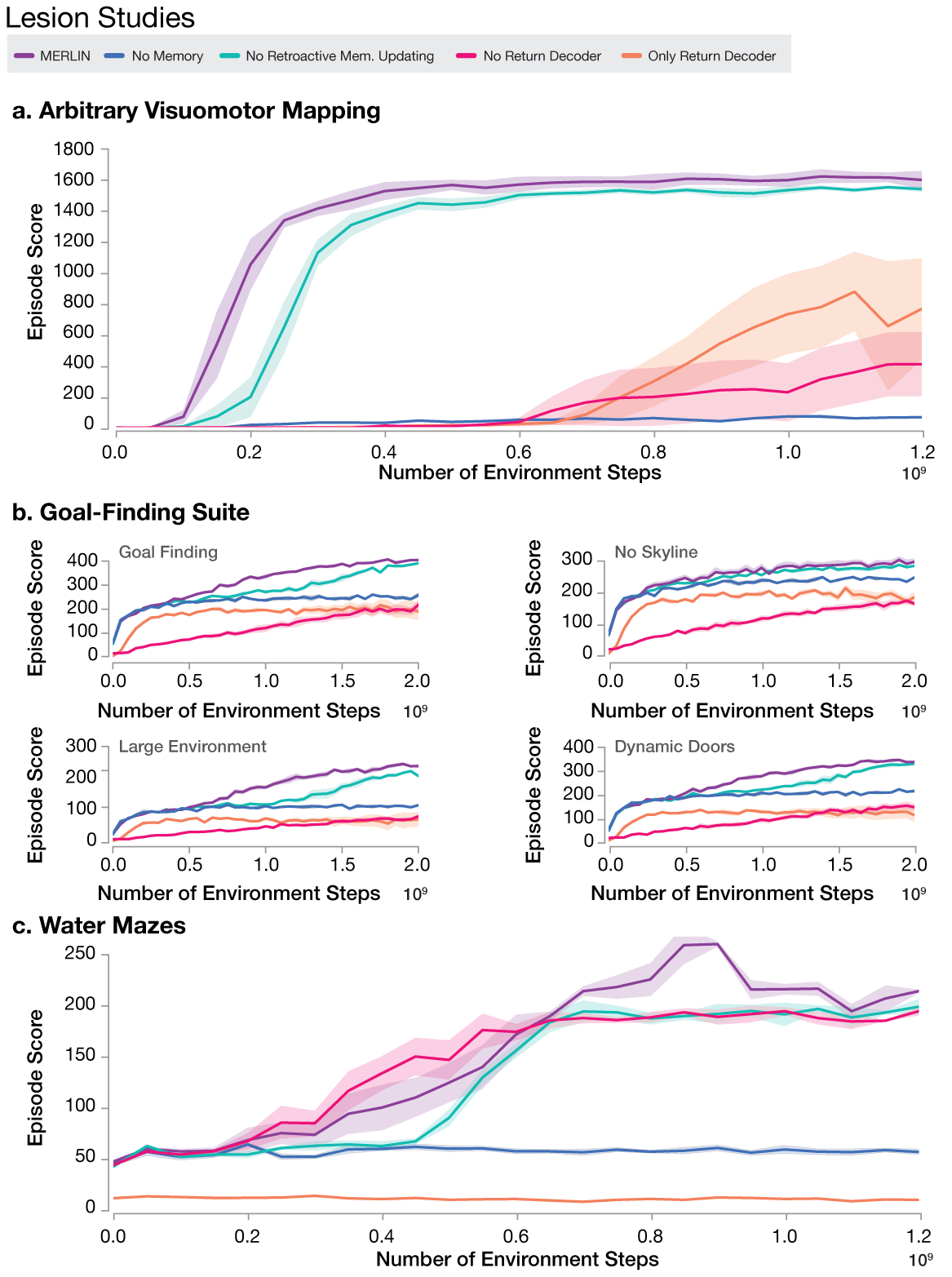}
\caption{\footnotesize{Lesion Studies for Various Tasks. \emph{No Memory}: The same as MERLIN but with no external memory and thus no reading from memory by either the memory-based predictor or the policy. \emph{No Retroactive Memory Updating}: There was no additional filtered sum of state variables placed in memory. \emph{No Return Decoder}: The cost term imposing that the latent state variable predict the return was removed. \emph{Only Return Decoder}: All memory-based predictor costs were removed except for the requirement to predict the return. \textbf{a-c}. Across all tested tasks, the lesions disrupted MERLIN's asymptotic performance or learning speed. The least harmful lesion was the removal of the retroactive memory updating, which modestly lowered final performance but primarily increased the time to learn. On the other hand, MERLIN was critically sensitive to the presence of its memory and the joint decoding of the return (future reward) with the observation. For goal-finding tasks, the number of environment steps shown was the total number taken across all four tasks.}}
\end{figure}

\begin{figure}[h!]
\center
\includegraphics[width=1\textwidth]{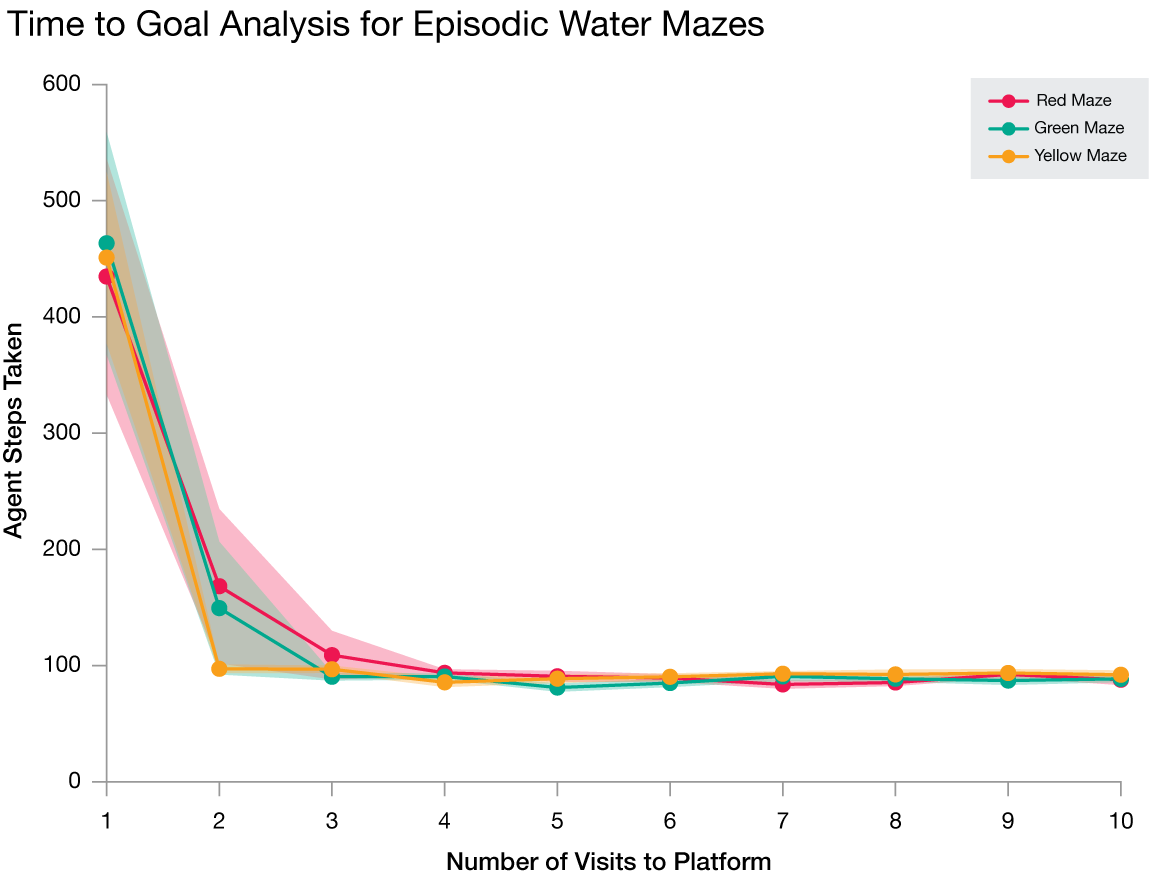}
\caption{\footnotesize{Time-to-Goal Analysis for Episodic Water Mazes. On the Episodic Water Mazes task, the time to find for MERLIN to find the platform decreased dramatically as a function of number of successful visits in each water maze colour.}}
\end{figure}

\begin{figure}[h!]
\center
\includegraphics[width=1\textwidth]{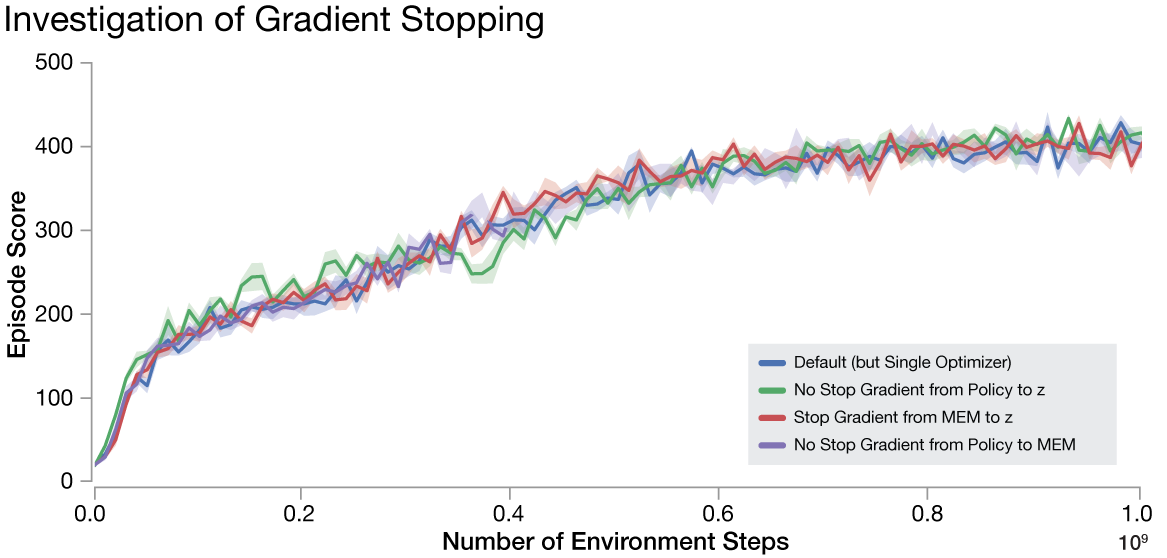}
\caption{\footnotesize{Investigation of Gradient Stopping. Eliminating or introducing the loss gradient from the state variable policy gradient to the state variable $z$, from the memory to the state variables that are inserted into it, or from the policy back to the memory did not obviously affect the performance of the model as shown here on the Goal Finding Task.}}
\end{figure}

\begin{figure}[h!]
\center
\includegraphics[width=1\textwidth]{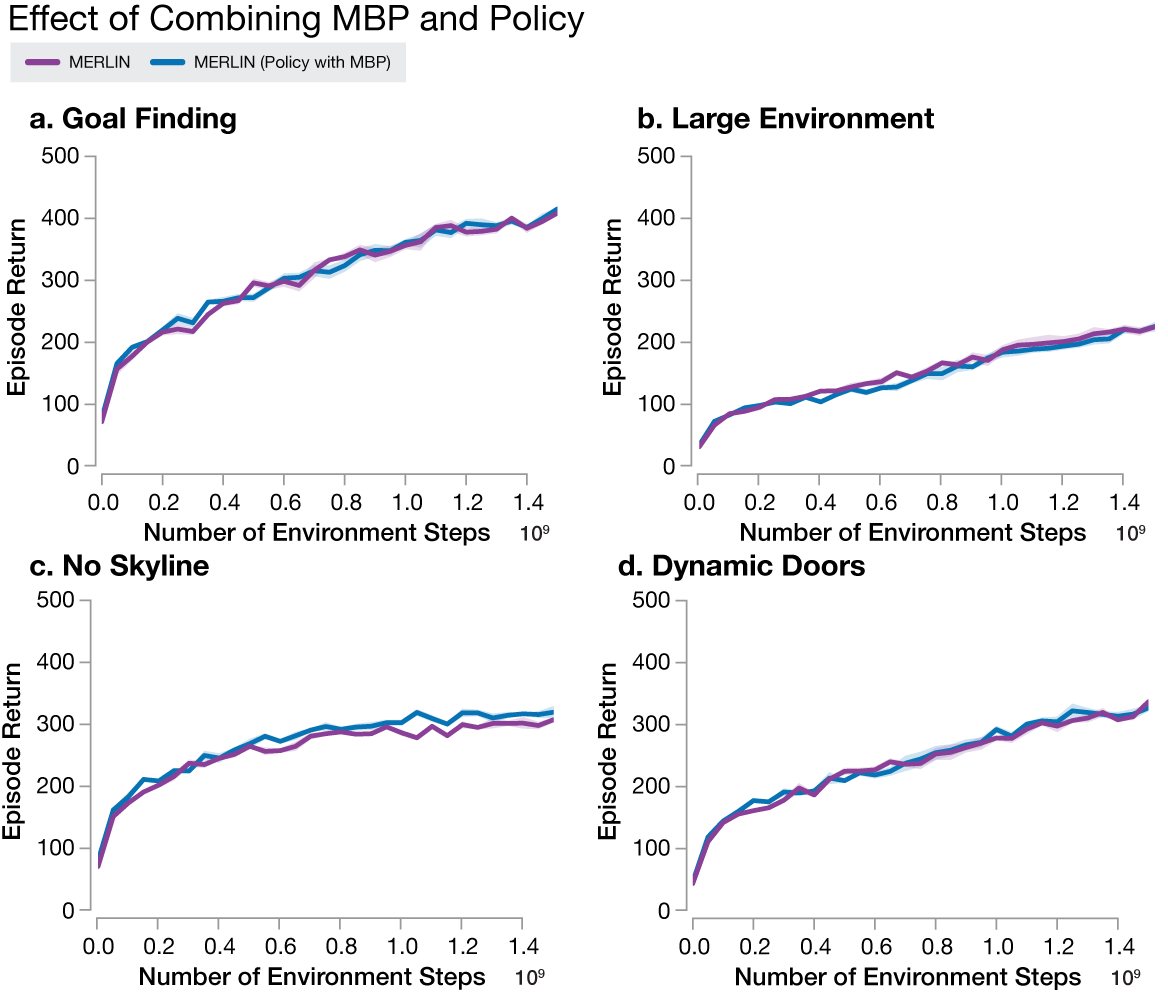}
\caption{\footnotesize{Effect of Combining MBP and Policy. The policy was combined with the MBP by creating a policy multi-layer perceptron with 200 hidden units whose input is the concatenation of the state variable, MBP output, and the MBP memory reads $[z_t,h_t,m_t]$. This version functions similarly, though the default MERLIN architecture better highlights the independence of predictive process from policy learning. A guiding purpose of our research program was to understand how memories can be encoded at long intervals before they are read out and accessed for behavioural decisions. Implementing the gradient blocks between the memory-based predictor network and the policy network was not a measure to improve performance, but rather it was an experimental choice to understand if it is truly necessary to implement end-to-end learning with gradients from decision events back to memory storage events. We consider our results to be very positive. Including or excluding these gradients made very little difference to agent performance, and this implies that we have built a memory system that works in a different way from other memory systems. (It does not use policy gradients.)}}
\end{figure}

\begin{figure}[h!]
\center
\includegraphics[width=1\textwidth]{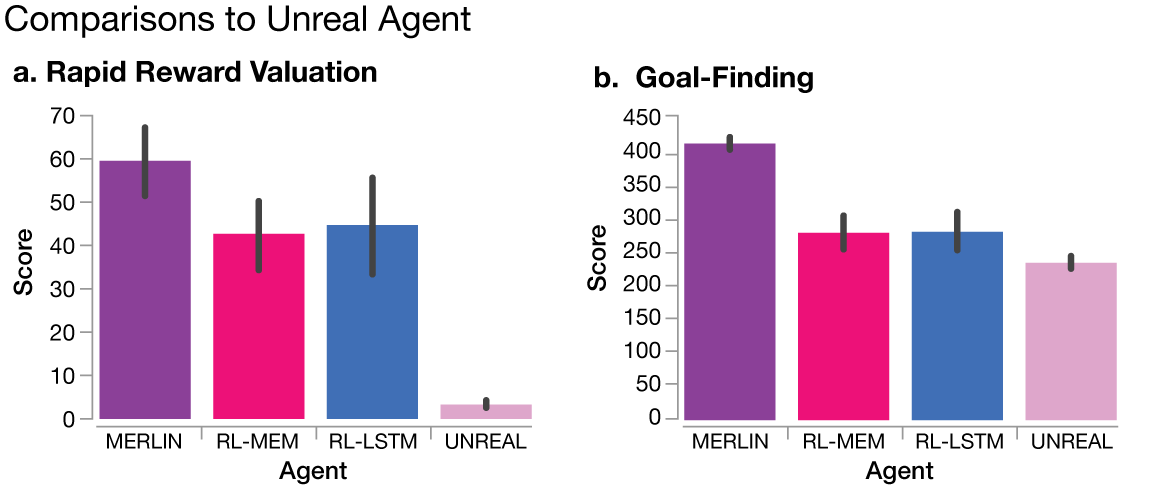}
\caption{\footnotesize{Comparisons to UNREAL Agent. We additionally compared the models against the UNREAL agent\cite{jaderberg2016reinforcement}, which is a state-of-the-art agent with auxiliary predictions provided by the authors. This model performed worse than MERLIN and the comparison models.}}
\end{figure}

\begin{figure}[h!]
\center
\includegraphics[width=0.8\textwidth]{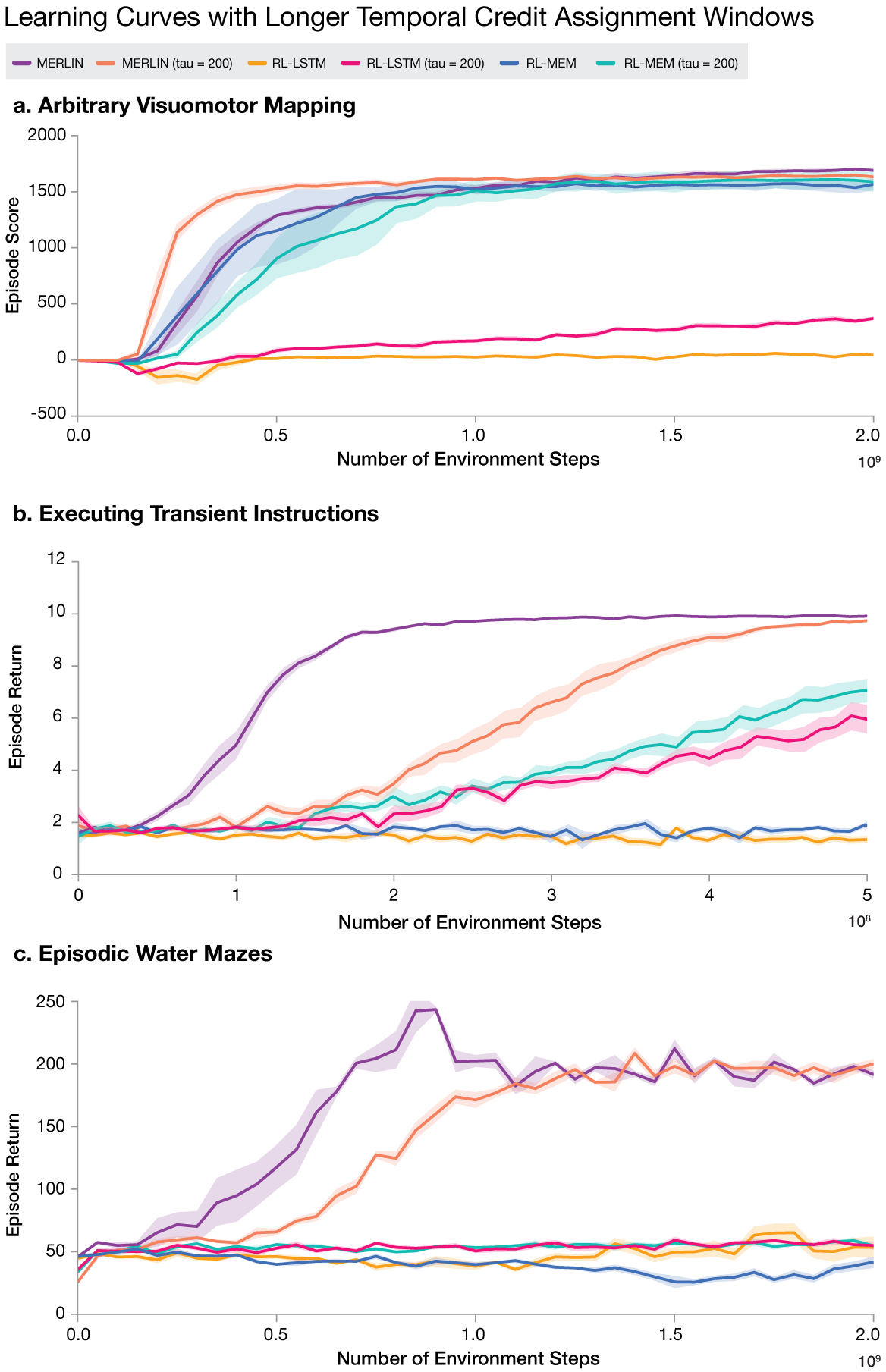}
\caption{\footnotesize{Learning Curves with Longer Temporal Credit Assignment Windows. For the task battery, setting $\tau$ to larger values for longer duration backpropagation lifted performance for the policy gradient RL models (RL-LSTM and RL-MEM), but they remained lower than MERLIN's. Additionally, for the task with intervening intervals between one performance phase and the next (Episodic Water Mazes), where one phase is in one maze and the next is in another with an initially long time between recurrences of the mazes, increasing the backpropagation through time $\tau$ did not help at all. In general, policy gradient RL did not learn to store essential information in memory when there was a delay between relevant observations and decisions that was longer than $\tau$.}}
\end{figure}

\begin{figure}[h!]
\center
\includegraphics[width=0.9\textwidth]{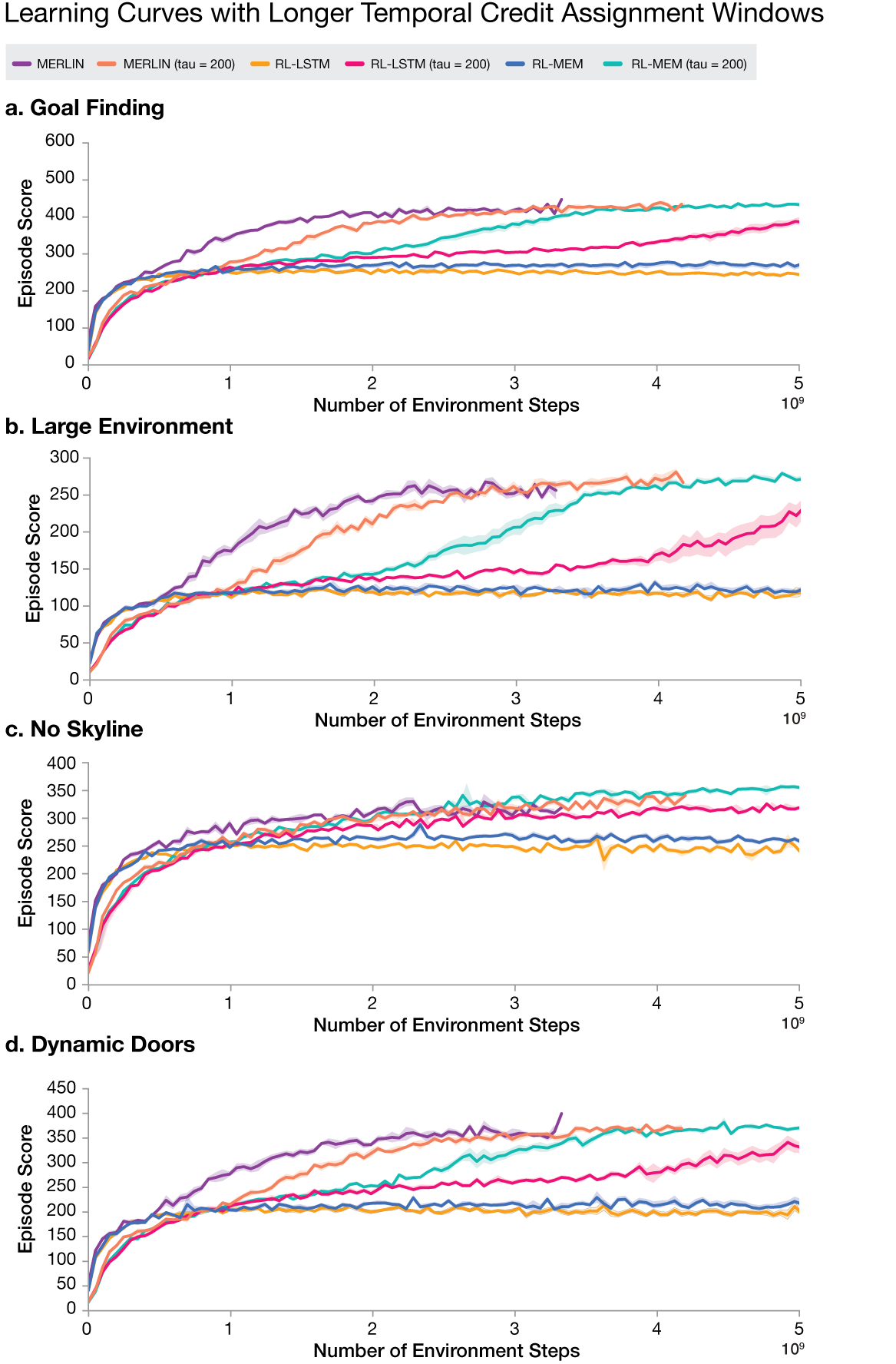}
\caption{\footnotesize{Learning Curves with Longer Temporal Credit Assignment Windows. For navigation tasks, setting $\tau$ to larger values for longer duration backpropagation lifted performance for the policy gradient RL models (RL-LSTM and RL-MEM), but they remained lower than MERLIN's. See Extended Figure 10 for further discussion.}}
\end{figure}

\begin{figure}[h!]
\center
\includegraphics[width=1\textwidth]{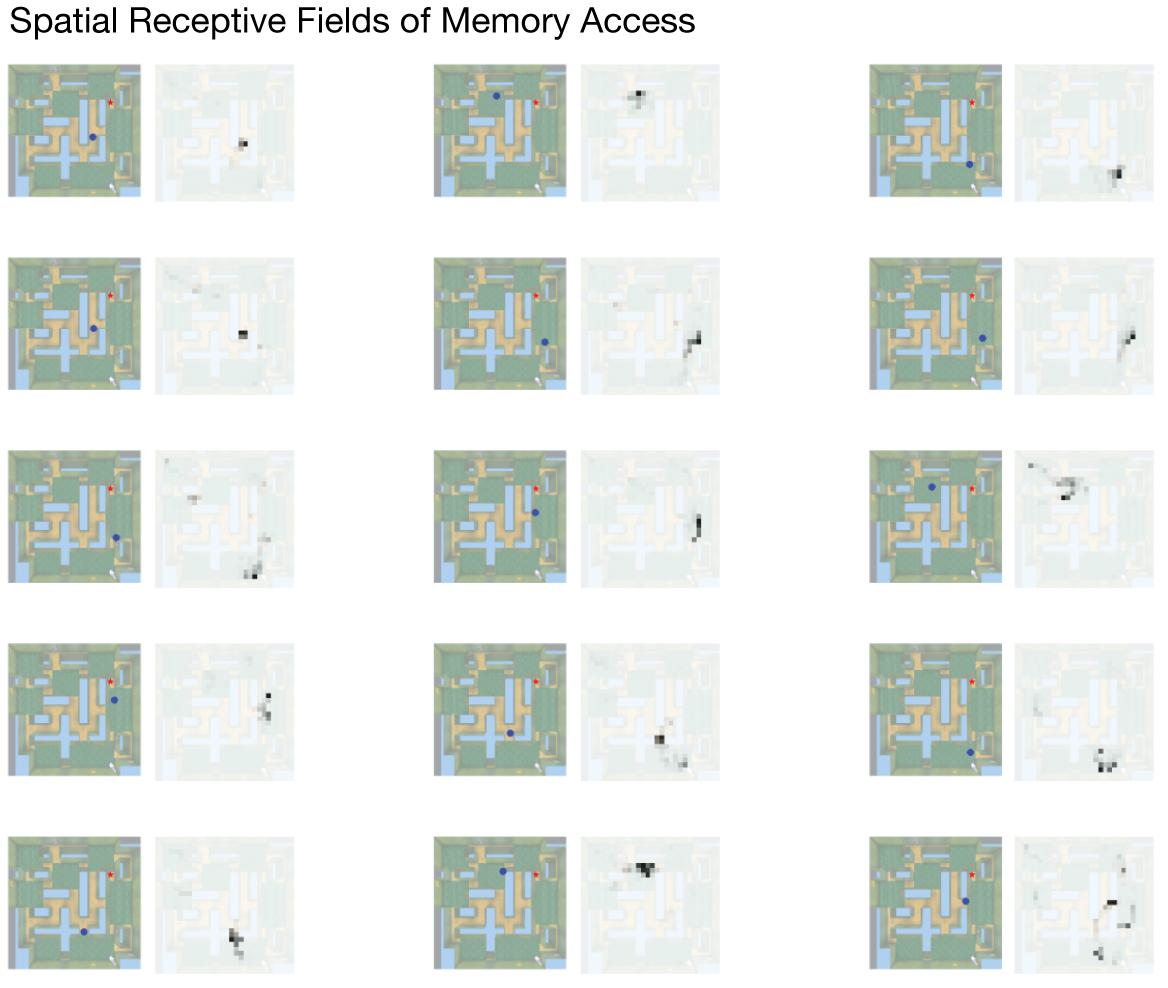}
\caption{\footnotesize{In one environment, we examined how much a MBP read attended to each memory row. \emph{Overhead View}: the agent position at the time of writing a first memory is shown by the blue dot. The goal position at the time of analysis is shown by the red dot. \emph{Transparent, Grey Overlay}: We calculated the spatial receptive field of the MBP's memory read access of that row. Most memory rows were read near to the place in which the rows were first written. See Methods Sec.~9.11.}}
\end{figure}

\begin{table}[h!]
\centering
\footnotesize{
 \begin{tabular}{||c c c c c||} 
 \hline
 & Goal-Finding & No Sky & Large Env. & Dynamic Doors \\ [0.5ex] 
 \hline\hline
 MERLIN & $\mathbf{40 \pm 1.1}$ & $\mathbf{29 \pm 1.4}$  & $\mathbf{24 \pm 1.3}$ & $\mathbf{32 \pm 0.8}$ \\ 
 RL-LSTM & $24 \pm 1.0$ & $24 \pm 0.7$ & $11 \pm 0.6$ & $19 \pm 1.2$ \\
 RL-MEM & $24 \pm 0.9$& $24 \pm 1.2$ & $11 \pm 1.1$ & $19 \pm 0.9$ \\
 Human & $31$ & $\mathbf{29}$ & $19$ & $24$ \\ [1ex] 
 \hline
 \end{tabular}}
 \caption{\footnotesize{Number of visits to goal in 90 seconds on goal-finding tasks with standard deviations across 10 replicated training runs with different random seeds. Bolded numbers are high scores. MERLIN's performance was at or above human-level after joint training across tasks.}}
 \label{table:multi}
\end{table}

\end{document}